\documentclass[conference]{IEEEtran}
\IEEEoverridecommandlockouts

\usepackage[utf8]{inputenc}
\long\def\ignoreme#1{}
\usepackage{graphicx}
\usepackage{array}
\usepackage{longtable}
\usepackage{xtab,booktabs}
\usepackage{subcaption}
\usepackage{algorithm}
\usepackage{algorithmicx}
\usepackage{algcompatible}
\usepackage{mathtools}
\usepackage[noend]{algpseudocode}
\algnewcommand\algorithmicforeach{\textbf{for each}}
\algdef{S}[FOR]{ForEach}[1]{\algorithmicforeach\ #1\ \algorithmicdo}

\newcommand\ms[1]{\mathsf{#1}}
\newcommand\mi[1]{\mathit{#1}}
\newcommand\mc[1]{\mathcal{#1}}
\newcommand\inv{\mi{Inv}}
\newcommand\relu{\ms{ReLU}}
\newcommand\softmax{\ms{softmax}}
\newcommand\feeds{\mi{feeds}}
\newcommand\supp{\mi{supp}}

\newcommand\mydef{\Coloneqq}
\newcommand\neurons{\mathcal{N}}

\newcommand\dpattern{\sigma}

\newcommand\on{\mi{on}}
\newcommand\off{\mi{off}}
\newcommand\decproc{\mi{DP}}
\newcommand\partialfun\rightharpoonup
\newcommand{\stitle}[1]{\vspace{3pt} \noindent\textbf{#1.}\
}
\newcommand\D{\mc{D}}
\renewcommand{\paragraph}[1]{\vspace{3pt} \noindent\textit{#1}\
}
\newcommand\expectation{\mathsf{E}}

\newtheorem{proposition}{Proposition}
\newtheorem{proof}{Proof}
\newtheorem{theorem}{Theorem}
\newtheorem{definition}{Definition}
\begin{document}


\title{Property Inference for Deep Neural Networks}
\author{\IEEEauthorblockN{Divya Gopinath\IEEEauthorrefmark{1},
Hayes Converse\IEEEauthorrefmark{2},
Corina S. P\u{a}s\u{a}reanu\IEEEauthorrefmark{1} and
Ankur Taly\IEEEauthorrefmark{3}}
\IEEEauthorblockA{\IEEEauthorrefmark{1} Carnegie Mellon University and NASA Ames \\
Email: divgml@gmail.com,corina.pasareanu@west.cmu.edu}
\IEEEauthorblockA{\IEEEauthorrefmark{2} University of Texas at Austin\\
Email: hayesconverse@gmail.com}
\IEEEauthorblockA{\IEEEauthorrefmark{3}Google AI\\
Email: ankur.taly@gmail.com}}

\maketitle


\begin{abstract}
We present techniques for {\em automatically} inferring formal properties of feed-forward neural networks.
We observe that
a  significant part (if not all) of the logic of feed forward networks
is captured in the activation status ($\on$ or $\off$) of its neurons.
We propose to extract patterns based on neuron decisions as preconditions that imply certain desirable output property e.g., the prediction being a certain class.
We present techniques to extract {\em input properties}, encoding convex predicates on the input space that imply given output properties and {\em layer properties}, representing network properties captured in the hidden layers that imply the desired output behavior. We apply our techniques on networks for the MNIST and ACASXU applications. Our experiments highlight the use of the inferred properties in a variety of tasks, such as explaining predictions, providing robustness guarantees, simplifying proofs, and network distillation.

\emph{Errata: This version updates~\cite{ase19} by correcting the definition of the three properties that were checked for ACASXU in Section~\ref{sec:acasxu}.
}
\end{abstract}


\section{Introduction}

%
%
Deep Neural Networks (DNNs) have emerged as a powerful mechanism for solving complex computational tasks, achieving impressive results that equal and sometimes even surpass human ability in performing these tasks.
%
%
However, the increased use of DNNs also brings along several safety and security concerns. These are due to many factors, among them {\em lack of robustness}. For instance, it is well known that DNNs, including highly trained and smooth networks, are vulnerable to adversarial perturbations. Small (imperceptible) changes to an input lead to misclassifications.
If such a classifier is used in the perception module of an autonomous car, the network’s decision on an adversarial image can have disastrous consequences.
DNNs also suffer from a {\em lack of explainability}: it is not well understood why a network
makes a certain prediction, which impedes on applications of DNNs
in safety-critical domains such as autonomous driving, banking, or medicine.
Finally, rigorous reasoning is obstructed by a lack of {\em intent} when designing neural networks, which only learn from examples, often without a high-level requirements specification. Such specifications are commonly used
when designing more traditional safety-critical software systems.


In this paper, we present techniques for {\em automatically} inferring formal properties of feed-forward neural networks. These properties are of the form $Pre \Rightarrow Post$.
$Post$ is a postcondition stating the desired output behaviour, for instance, the network's prediction being a certain class.
$Pre$ is a precondition that we automatically infer and can serve as a {\em formal explanation} for why the output property holds.
We study {\em input properties} which encode predicates in the input space that imply a given output property
We further study {\em layer properties} which group inputs that have common characteristics observed at an intermediate layer and that together imply the desired output behavior
The intention is to capture properties based on the {\em features} extracted by the network.

There are many choices for defining network properties that are appropriate preconditions for network behavior.  In this work, we infer properties corresponding to {\em decision patterns} of neurons in the DNN. Such patterns prescribe which neurons are $\on$ or $\off$ in various layers. For neurons implementing the $\relu$ activation function, this amounts to whether the neuron output is greater than zero ($\on$) or equal to zero ($\off$). We focus on these simple patterns because they are easy to compute and have simple mathematical representations. Furthermore, they define natural partitions on the input space, grouping together inputs that are processed the same by the network and that yield the same output. Other obvious, more complex properties (e.g. use a positive threshold rather than zero for the activation functions, use linear  combinations on neuron values) are left for study in future work.

We define input properties based on patterns that constrain the activation status ($\on$ or $\off$) of all neurons up to an intermediate layer. Such patterns form convex predicates in the input space. Convexity is attractive as it makes the inferred properties easy to visualize and interpret. Furthermore, convex predicates can be solved efficiently with existing linear programming solvers. Analogously, we define layer properties based on patterns that constrain the activation status at an intermediate layer. Layer patterns define convex regions over the values at an intermediate layer and can be expressed as unions of convex regions in the input space.

Another motivation for studying decision patterns is that they are analogous to path constrains in program analysis. Different program paths capture different input-output behaviour of the program. Similarly, different neuron decision patterns capture different behaviours of a DNN. It is our proposition that we should be able to extract succinct input-output properties based on decision patterns that together explain the behavior of the network, %
and can act as formal specifications of networks.
We present two techniques to extract network properties. Our first technique is based on iteratively refining decision patterns while leveraging an off-the-shelf decision procedure. We make use of the decision procedure Reluplex~\cite{KaBaDiJuKo17Reluplex}, designed to prove properties of feed-forward ReLU networks, but other decision procedures can be used as well. Our second technique uses decision tree learning to directly {\em learn} layer patterns from data. The learned patterns can be formally checked using a decision procedure. In lieu of a formal check, which is typically expensive, one could empirically validate the learned patterns over a held-out dataset to obtain confidence in their precision.

We consider this work as a first step in the study of formal properties of DNNs. As a proof of concept, we present several different applications. We learn input and layer properties for an MNIST network, and demonstrate their use in providing robustness guarantees, explaining the network's decisions and debugging misclassifications made by the network. We  also  study  the  use  of  patterns  at intermediate  layers  as  interpolants  in  the  proof  of  given input-output  properties  for a network modeling a safety-critical system for unmanned aircraft control (ACAS XU)~\cite{ACASXU}.
The learned patterns help decompose the proofs thereby making them computationally efficient. 
Finally, we discuss a somewhat radical application of the learned patterns in distilling~\cite{teacher-student} the behavior of DNNs. The key idea is to use the patterns that have high support as distillation rules that directly determine the network's prediction without evaluating the entire network. This results in a significant speedup without much loss of accuracy.

\section{Background}
\label{sec:background}
A neural network defines a function $F: \rm I\!R^n \rightarrow \rm I\!R^m$ mapping an input vector of real values $X \in \rm I\!R^n$ to an output vector $Y \in \rm I\!R^m$. For a classification network, the output defines a score (or probability) across $m$ classes, and the class with the highest score is typically the predicted class. A \emph{feed forward} network is organized as a sequence of  layers with the first layer being the input. 
Each intermediate layer consists of computation units called \emph{neurons}. Each neuron consumes a linear combination of the outputs of neurons in the previous layer, applies a non-linear activation function to it,  and propagates the output to the next layer. The output vector $Y$ is a linear combination of the outputs of neurons in the final layer.
For instance, in a Rectified Linear Unit (ReLU) network, each neuron applies the activation function $\relu(x) = max(0,x)$.
Thus, the output of each neuron is of the form $\relu(w_1\cdot v_1 + \ldots + w_p\cdot v_p + b)$ where $v_1, \ldots v_p$ are the outputs of the neurons from the previous layer, $w_1, \ldots, w_p$ are the weight parameters, and $b$ is the bias parameter of the neuron.\footnote{Most classification networks based on ReLUs typically apply a softmax function at the output layer to convert the output to a probability distribution. We express such networks as $F \mydef = \softmax(G)$, where $G$ is a pure ReLU network, and then focus our analysis on the network $G$. Any property of the output of $F$ is translated to a corresponding property of $G$.}

\stitle{Example}
\ignoreme{
\begin{figure}[t]
\begin{center}
\includegraphics[scale=0.2]{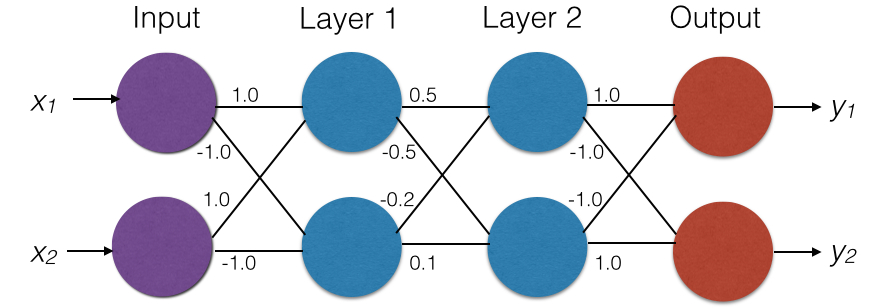}
\end{center}
\caption {Example \label{fig:example}}
\end{figure}
}
\begin{figure*}[t]
\begin{subfigure}[c]{0.5\textwidth}
\begin{center}
\includegraphics[scale=0.2]{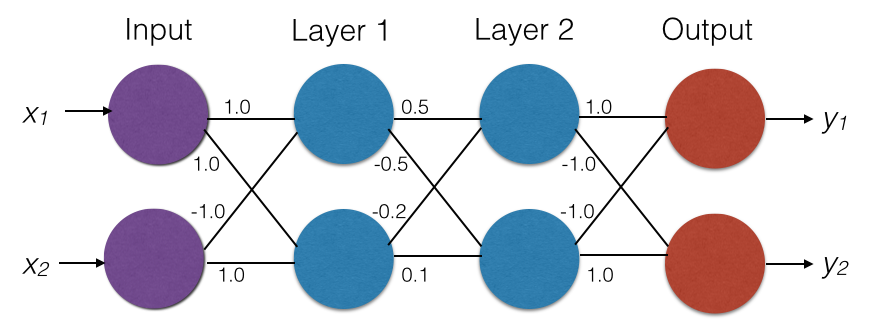}
\caption{Example \label{fig:example}}
 \end{center}
\end{subfigure}
\hfill
\begin{subfigure}[c]{0.5\textwidth}
\begin{center}
\includegraphics[scale=0.3]{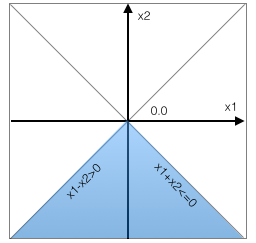}
\caption{Input property for prediction ``1" \label{fig:invariants}}
\end{center}
\end{subfigure}
\caption{Example neural network and input contract}
\vspace{-0.7cm}
\end{figure*}
We use a simple feed forward ReLU network, shown in Figure~\ref{fig:example}, as a running example throughout this paper. The network has four layers: one input layer, two hidden layers and one output layer. It takes as input a vector of size 2. The output vector is also of size 2, indicating classification scores for 2 classes. All neurons in the hidden layers use the ReLU activation function. The final output is a linear combination of the outputs of the neurons in the last hidden layer. Weights are written on the edges. For simplicity, all biases are zero.
Consider the input $[1.0,-1.0]$. The output on this input is $F([1.0,-1.0])= [y_1, y_2] = [1.0,-1.0]$. To see this, notice that the output of the first hidden layer is $[v_{1,1}, v_{1, 2}]=[\relu(1.0\cdot1.0-1.0\cdot-1.0),\relu(1.0\cdot1.0+1.0\cdot-1.0)]=[2.0,0.0]$. This feeds into the second hidden layer whose output then is  $[v_{2, 1}, v_{2, 2}]=[\relu(0.5\cdot2.0-0.2\cdot0.0),\relu(-0.5\cdot2.0+0.1\cdot0.0)]=[1.0, 0.0]$. This in turn feeds into the output layer which computes $[y_1, y_2]=[1.0\cdot1.0-1.0\cdot0.0, -1.0\cdot1.0+1.0\cdot0.0] = [1.0, -1.0]$.



A feed forward network is called \emph{fully connected} if all neurons in a hidden layer feed into all neurons in the next layer; the network in Figure~\ref{fig:example} is such a network. Convolutional Neural Networks (CNNs) are similar to ReLU networks, but in addition to (fully connected) layers, they may also contain {\em convolutional layers} which compute multiple convolutions of the input with different filters and then apply the ReLU activation function.
For simplicity, we focus our discussion on ReLU networks, but our work 
applies to all piece-wise linear networks, including ReLUs and CNNs (and in experiments we describe an analysis for a CNN).

\stitle{Notations and Definitions}
All subsequent notations and definitions are for a feed forward ReLU network $F$, often referred to implicitly. We use uppercase letters to denote vectors and functions, and lowercase letters for scalars.
We use $N, N', N_1, \ldots$ to range over neurons, and $\neurons$ for the set of all neurons in the network.
For any two neurons $N_1, N_2$, the relation $N_1 \prec N_2$ holds if and only if the output of neuron $N_1$ feeds into neuron $N_2$, either directly or via intermediate layers.
We define $\feeds(N) \mydef \{N'~\vert~N' \prec N\}$, and extend it to sets of neurons in the natural way.

The output of each neuron $N$ can be expressed as a function of the input $X$. We abuse notation and use $N(X)$ to denote this function. It is defined recursively via neurons in the preceding layer. That is, if $N_1, \ldots, N_p$ are neurons from the preceding layer that directly feed into $N$, then $N(X) = \relu(w_1\cdot N_1(X) + \ldots + w_2\cdot N_2(X) + b)$.
For ReLU networks, $N(X)$ is always greater than or equal to $0$. We say that the neuron is $\off$ if $N(X) = 0$ and $\on$ if $N(X) > 0$. This essentially splits the cases when the ReLU fires and does not fire. As we will see in Section~\ref{sec:theory}, the $\on$/$\off$ activation status of neurons is our key building block for defining network properties.

\section{Network Properties}\label{sec:theory}



Our goal is to extract succinct input-output characterizations of the network behaviour, that can act as formal specifications for the network. The network itself provides an input-output mapping but of course this is uninteresting. Ideally we should group together inputs that lead to the same output and express that in concise mathematical form.
To this end we propose to infer {\em input properties} wrt a given output property $P$.
An input property is a predicate over the input space, such that, all inputs satisfying it evaluate to an output satisfying the property $P$. In other words, an input property is a {\em precondition} for {\em postcondition} $P$. Together, the input property and the post condition form a formal {\em contract} for the network.
An example of an output property for a classification network is that the top predicted class is $c$, i.e., $P(Y) \mydef argmax(Y) = c$.
Such properties are called \emph{prediction postconditions}.

In this work, we infer \emph{input properties} that characterize inputs that are processed in the same way by the network, i.e. they follow the same on/off activation pattern up to some layer
and  define convex regions in the input space. There may be many such convex regions for a particular output property (say a particular prediction). The union of these regions fully captures the behavior of the network wrt the output property. In practice it may be too expensive to compute precisely this union but we show that even computing a subset of these regions can be useful for many applications.

We further study \emph{layer properties} which encode common properties at an intermediate layer that imply the desired output behavior. Neural networks work by applying layer after layer of transformations over the inputs, to extract {\em important features} of the data, and then make decisions based on these features. Thus layer properties can potentially capture common characteristics over the extracted features, allowing us to get insights into the inner workings of the network.
Similar to input properties, we seek to infer layer properties by studying the activation patterns of the network.
Unlike input properties, layer properties do not map to convex regions in the input space, but rather to unions of convex input regions.




\stitle{Decision Patterns}
We infer network properties based on \emph{decision patterns} of neurons in the network.
A decision pattern $\dpattern$ specifies an activation status ($\on$ or $\off$) for some subset of neurons. All other neurons are don't care. We formalize decision patterns $\dpattern$ as partial functions $\neurons \partialfun \{\on, \off\}$, and write $\on(\dpattern)$ for the set of neurons marked $\on$, and $\off(\dpattern)$ be the set of neurons marked $\off$ in the pattern $\dpattern$.
Each decision pattern $\dpattern$ defines a predicate $\dpattern(X)$ that is satisfied by all inputs whose evaluation achieves the same activation status for all neurons as prescribed by the pattern.
\begin{equation}\label{eqn:dpattern}
\dpattern(X) \mydef \bigwedge_{N \in \on(\dpattern)} N(X) > 0 ~ \wedge ~ \bigwedge_{N \in \off(\dpattern)} N(X) = 0
\end{equation}
A decision pattern $\dpattern$ is a network property wrt a postcondition $P$ if:
\begin{equation}\label{eqn:defprop}
\forall X: \dpattern(X) \implies P(F(X)).
\end{equation}


We seek \emph{minimal} patterns $\dpattern$ which have the property that dropping (which amounts to unconstraining) any neuron from the pattern invalidates it. Minimality helps in getting rid of unnecessary constraints, and ensuring that more inputs can satisfy the property.

The {\em support} of a pattern, denoted by $\supp(\dpattern)$, is a measure of the number of inputs that follow the pattern. Formally, it is the total probability mass of inputs satisfying $\dpattern$, under a given input distribution. In the absence of an explicit input distribution, support can be measured empirically based on a training or test dataset. For large networks a formal proof for $\forall X: \dpattern(X) \implies P(F(X))$ may not be feasible. In such cases, one could aim for a probabilistic guarantee that the conditional expectation (denoted $\expectation$) of $P(F(X))$ given $\dpattern(X)$ is above a certain threshold, i.e., $\expectation(P(F(X))~\vert~\dpattern(X)) \geq \tau$.\footnote{This is similar to the probabilistic guarantee associated with ``Anchors''~\cite{Ribeiro0G18},
which we discuss further in Section~\ref{sec:related}.}

\subsection{Input Properties}
To build input properties we infer input properties that are convex predicates in the input space implying a given postcondition.
Given that feed forward ReLU networks encode highly non-convex functions, the existence of input properties is itself interesting.
To identify input properties, we consider decision patterns wherein for each neuron $N$ in the pattern, all neurons that feed into $N$ are also included in the pattern. We call such patterns $\prec$-closed. We show that $\prec$-closed patterns capture convex predicates in the input space.
\begin{theorem}\label{thm:convexity}
For all $\prec$-closed patterns $\dpattern$, $\dpattern(X)$ is convex, and has the form:
\begin{equation*}
\bigwedge_{i \, in \, 1 .. \vert\on(\dpattern)\vert} W_i\cdot X + b_i > 0 ~ \wedge ~ \bigwedge_{j \, in \, 1 .. \vert\off(\dpattern)\vert} W_j\cdot X + b_j \leq 0
\end{equation*}
Here $W_i, b_i, W_j, b_j$ are some constants derived from the weight and bias parameters of the network.
\end{theorem}
The proof is provided in the Appendix. It is based on induction over the depth of neurons in the pattern $\dpattern$. It shows that the value of any neuron in the pattern can be expressed as a linear combination of the inputs and that each on/off activation adds a linear constraint to the input predicate.
\footnote{The theorem can also be proven by representing the network as a \emph{conditional affine transformation} as shown in~\cite{DBLP:conf/sp/GehrMDTCV18}.}
Thus, an input property can be obtained by identifying a  $\prec$-closed pattern $\dpattern$ such that $\forall X: \dpattern(X) \implies P(F(X))$. For convex postconditions $P$, we show that an input property can be identified using any input $X$ whose output satisfies $P$. For this, we consider the  \emph{activation signature} of $X$, which is a decision pattern $\dpattern_X$ that constrains the activation status of {\em all} neurons to that obtained during the evaluation of $X$.

\begin{definition}\label{def:activation_sig}
Given an input $X$, the activation signature of $X$ is a decision pattern $\dpattern_X$ such that for each neuron $N \in \neurons$, $\dpattern_X(N)$ is $\on$ if $N(X) > 0$, and $\off$ otherwise.
\end{definition}
It is easy to see that $\dpattern_X$ is a $\prec$-closed pattern.
Thus, following Theorem 1,  $\dpattern_X$ can be used to obtain an input property, i.e. a property that implies a desired output behavior.
We state this result as a proposition, which will be used in Section~\ref{sec:methods}.

\begin{proposition}\label{prop:activation_sig}
Given a convex postcondition $P$ and an input $X$ whose output satisfies $P$ (i.e., $P(F(X)$ holds), the
following holds. There exist parameters $W, b$ such that:
\begin{itemize}
    \item [(A)] $\forall X':\dpattern_X(X') \implies F(X') = W\cdot X' + b$
    \item [(B)] The predicate $\dpattern_X(X') ~\wedge~P(W\cdot X' + b)$ is an input property.
\end{itemize}
\end{proposition}

\stitle{Example}
We illustrate input properties on the network shown in Figure~\ref{fig:example} (introduced in Section~\ref{sec:background}). Consider the postcondition that the top prediction is class $1$, i.e., $P([y_1, y_2]) \mydef y_1 > y_2$. Let $N_{1,1}, N_{1,2}$ be the neurons in the first hidden layer, and $N_{2,1}, N_{2,2}$ be the neurons in the second hidden layer. Consider the pattern $\dpattern = \{N_{1,1} \rightarrow \on, N_{1,2} \rightarrow \off\}$. We argue that this pattern is an input property wrt $P$.
Since $N_{1,1}$ is $\on$ it must be the case that the values that feed into $N_{1,1}$ (which have the form $x_1-x_2$) are positive, hence the inputs satisfy $x_1-x_2>0$.
Furthermore, since $N_{1,2}$ is $\off$ it must be the case that the values that feed into $N_{1,2}$ (which have the form $x_1+x_2$) are negative, hence the inputs satisfy $x_1+x_2\leq0$.
Now notice that all the inputs that satisfy these two constraints also satisfy
neuron $N_{2,1}$ is always $\on$ and neuron $N_{2,2}$ is always $\off$.
This is because the value that feeds into $N_{2,1}$ is $0.5 \cdot (x_1 - x_2)$ which must be positive (since
$x_1-x_2>0$). Similarly the value that feeds into $N_{2,2}$ is $-0.5\cdot (x_1 - x_2)$ which must be negative. Consequently the output  $[y_1, y_2] = [1.0\cdot N_{2,1}(X) - 1.0\cdot N_{2,2}(X), -1.0\cdot N_{2,1}(X) + 1.0\cdot N_{2,2}(X)]=[0.5\cdot(x_1-x_2),-0.5\cdot(x_1-x_2)]$ always satisfies $y_1 > y_2$ (when $x_1-x_2>0$),
making the pattern a precondition for the property $P$. The pattern is $\prec$-closed, and therefore by Theorem~\ref{thm:convexity}, the predicate $\dpattern(X)$ is convex.
The predicate $\dpattern(X) = N_{1,1}(X) > 0~\wedge~N_{1,2}(X) = 0$ (see Equation~\ref{eqn:dpattern}) amounts to the convex region $x_1 - x_2 > 0 \wedge x_1 + x_2 \leq 0$ (shown in blue in Figure~\ref{fig:invariants}) and is minimal. 

\ignoreme{
\begin{figure}[t]
\begin{center}
\includegraphics[scale=0.20]{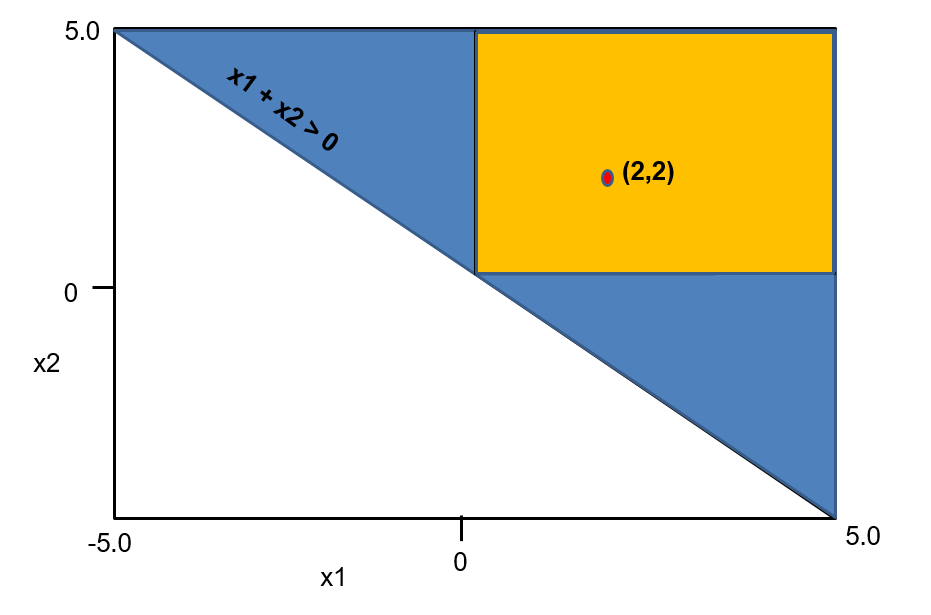}
\end{center}
\vspace{-0.5cm}
\caption{Input invariants for the network in Figure~\ref{fig:example}. \label{fig:invariants}}
\vspace{-0.5cm}
\end{figure}
}
\subsection{Layer Properties}
While inferred input properties may be easy to interpret, they often have tiny support. For instance, a property defined based on the activation signature of an input $X$ may only be satisfied by $X$, and possibly a few other inputs that are syntactically close to $X$. Ideally, we'd like properties to group together inputs that are semantically similar in the eye of the network. To this end, we focus on decision patterns at an intermediate layer that capture high-level features.

A layer property for a  postcondition $P$ encodes a decision pattern $\dpattern^l$ over neurons in a specific layer $l$ that satisfies $\forall X: \dpattern^l(X) \implies P(F(X))$.\footnote{For simplicity, we restrict ourselves to computing properties with respect to a single internal layer but the approach extends to multiple layers.}

Note that a layer property is convex in the space of values at that layer, but not in the input space. However, it is simple to express a layer property as a disjunction of input preconditions. This is achieved by extending a layer pattern with all possible patterns over neurons that feed into the layer (directly or indirectly). Each such extended pattern is $\prec$-closed, and therefore convex (by Theorem~\ref{thm:convexity}). We formulate this  connection between layer and input properties in the following proposition.
\begin{proposition}\label{prop:layerinvariant}
Let $\dpattern^l$ be a layer property for an output property $P$. Let $\neurons^l$ be the set of neurons constrained by $\dpattern^l$, and let $\dpattern_1, \ldots, \dpattern_p$ be all possible decision patterns over neurons in $\feeds(\neurons^l).$\footnote{There are two $2^{\vert\feeds(\neurons^l)\vert}$ such patterns.}
Then the following statements hold:

\begin{itemize}

    \item [(A)] For each $i$, $\dpattern^l(X)~\wedge~\dpattern_i(X)$ is an input property.
    \item [(B)] $\dpattern^l(X) \Leftrightarrow \bigvee_i (\dpattern^l(X)~\wedge~\dpattern_i(X))$.
\end{itemize}
\end{proposition}
Thus, layer properties can be seen as a grouping of several input properties as dictated by an internal layer.
We note that identifying the right layer is key here. For instance, if one picks a layer too close to the output then the layer property may span all possible input properties, which is uninteresting.
In general, the choice of layer would depend on the application. We discuss it further in Section~\ref{sec:applications}.

\stitle{Example}
Let us revisit the example in Figure~\ref{fig:example} for the postcondition that the top prediction is class $1$, i.e., $P([y_1, y_2]) \mydef y_1 > y_2$. A layer pattern for this property is $\{N_{2,1}\rightarrow \on, N_{2,2} \rightarrow \off\}$. It is easy to see
that for all inputs satisfying this pattern, the output $[y_1, y_2] = [1.0\cdot N_{2,1}(X) - 1.0\cdot N_{2,2}(X), -1.0\cdot N_{2,1}(X) + 1.0\cdot N_{2,2}(X)]$ will satisfy $y_1 > y_2$, making the pattern a layer property wrt $P$. \ignoreme{The pattern is also minimal as unconstraining either neuron $N_{2,1}$ or $N_{2,2}$ invalidates it as a layer property wrt $P$. } The pattern is satisfied by the input $[1.0, -1.0]$. The execution of this input involves neuron $N_{1,1}$ being $\on$ and neuron $N_{1,2}$ being $\off$. Consequently, by proposition~\ref{prop:layerinvariant} (part (A)), the extended pattern $\{N_{1,1} \rightarrow \on, N_{1,2} \rightarrow \off, N_{2,1}\rightarrow \on, N_{2,2} \rightarrow \off\}$ is an input property wrt $P$.

\subsection{Interpreting and Using Inferred Network Properties}\label{subsec:interpreting_invariants}

\stitle{Robustness guarantees and adversarial examples} We first remark that provably-correct input and layer properties defined wrt prediction postconditions characterize regions in the
input space in which the network is guaranteed to give the same label, i.e. the network is robust. Inputs generated from counter-examples of pattern candidates that fail to prove represent potential adversarial examples, as they are close (in the Euclidean space) to (regions of) inputs that are classified differently. Furthermore, they are semantically similar to benign ones (since they follow the same decision pattern) yet are classified differently.
We show such examples in Section~\ref{sec:applications}.

\stitle{Explaining network predictions}
Neural networks are infamous for being complex black-boxes~\cite{L16,DK17}. An important problem in interpreting them is to understand why the network makes a certain prediction on an input. Predictions properties (that ensure that the prediction is a certain class)
can be used to obtain such explanations. But, such properties are useful explanations only if they are themselves understandable.
Inferred input properties are useful in this respect as they trace convex regions in the input space. Such regions are easy to interpret when the input space is low dimensional.

For networks with high-dimensional inputs (e.g., image classification networks) input properties may be hard to interpret or visualize. The conventional approach here is to explain a prediction by assigning an importance score, called \emph{attribution}, to each input feature~\cite{SimonyanVZ13,SundararajanTY17}. The attributions can be visualized as a heatmap overlayed on the visualization of the input. In light of this, we propose two different methods to obtain similar
visualizations from input properties.
We note that in contrast to attributions, which help explain predictions for individual inputs, our proposed input properties help explain the predictions for regions of the input space. Furthermore, and in contrast to existing attribution methods, they provide formal guarantees as the computed explanations are themselves network properties that imply the given postcondition.

\paragraph{Under-approximation Boxes.} As stated in Theorem~\ref{thm:convexity}, an input property consists of a conjunction of linear inequations, which can be solved efficiently with existing Linear Programming (LP) solvers.
We propose computing {\em under-approximation boxes} (i.e. bounds on each dimension) as a way to interpret input properties.
Specifically, we use LP solving (after a suitable re-writing of the constraints)\footnote{We replace each occurrence of variable $x_i$ with $lo_i$ or $hi_i$ based on the sign of the coefficient in the inequalities. See the Appendix for details on the computation of under-approximation boxes.} to find solution intervals $[lo_i,hi_i]$ for each input dimension $i$ such that $\sum_i(hi_i-lo_i)$ is maximized. As there are many such boxes, we constrain each box to include as many inputs from the support as possible.
These boxes provide simple mathematical representations of the properties, and are easy to visualize and interpret. Note that the under-approximating boxes are themselves network properties that formally imply the input properties and hence the given postcondition.

\paragraph{Minimal Assignments.} We also propose another natural way to interpret both input and layer properties through the lens of a particular input. Analogous to attribution methods, we aim to determine which input dimensions (or features) are most relevant for the satisfaction of the property. Every concrete input defines an assignment to the input variables $x_1=v_1 \wedge x_2=v_2 \wedge .. \wedge x_n=v_n$ that satisfies $\sigma(X)$. The problem now is to find a {\em minimal assignment} that still leads to the satisfaction of the property, i.e., a minimal subset of the assignments such that $x_{k_1}=v_{k_1} \wedge x_{k_2}=v_{k_2} \wedge .. \wedge x_{k_n}= v_{k_n} \implies \sigma(X)$. The problem has been studied in the constraint solving literature, and is known to be computationally expensive~\cite{Dillig:2012:MSA:2362216.2362255}. 
We adopt a greedy approach that eliminates constraints iteratively and stops when $\sigma(X)$ is no longer implied; the checks are performed with a decision procedure. The resulting constraints are also network properties that formally guarantee the corresponding postcondition.

\stitle{Layer Patterns as Interpolants}
For deep networks deployed in safety-critical contexts, one often wishes to a prove a contract of the form $A \implies B$, which says that for all inputs $X$ satisfying $A(X)$, the corresponding output $Y$ ($= F(X)$) satisfies $B(Y)$.
For the ACASXU application, there are several desirable properties of this form, wherein, $A$ is a set of constraints defining a single or disjoint convex regions in the input space, and $B$ is an expected output advisory.
Formally, proving such properties for multi-layer feed forward networks is computationally expensive~\cite{KaBaDiJuKo17Reluplex}. 
We show that the inferred network patterns, in particular layer patterns, help decompose proofs of such properties by serving as useful interpolants~\cite{interpolant_mcmillan}.
Given a layer pattern $\dpattern^l$, we propose the following rule to decompose a proof.
\begin{equation}\label{eqn:proof_decomposition}
\dfrac{(A \implies \dpattern^l), (\dpattern^l \implies B)}{(A \implies B)}
\end{equation}
%
%
Thus, to prove $A \implies B$, we must first identify a layer pattern $\dpattern^l$ that implies output property $B$, and then attempt the proof $A \implies \dpattern^l$ on the smaller network up to layer $l$.
Additionally, once a layer pattern $\dpattern^l$ is identified for a property $B$, it can be reused to prove other properties involving $B$. In Section~\ref{sec:applications}, we show that this decomposition leads to significant savings in verification time for properties of the ACASXU network.

\stitle{Distilling rules from networks}
Distillation is the process of approximating the behavior of a large, complex deep network with a smaller network~\cite{teacher-student}. The smaller network is meant to be favorable to deployment under latency and compute constraints while having comparable accuracy.  We show that layer patterns with high support provide a novel way to perform such distillation. Suppose $\dpattern^l$ is a pattern at an intermediate layer $l$ that implies that the prediction is a certain class $c$. For any input $X$, we can execute the network up to layer $l$, and check if the activation statuses of the neurons in layer $l$ satisfy the pattern $\dpattern^l$. If they do then we can directly return the prediction class $c$. Otherwise we continue executing the network. Thus for all inputs where the pattern is satisfied, we replace the cost of executing the network from layer $l$ onward (possibly involving several matrix multiplications) with simply checking the pattern $\dpattern^l$. The savings could be substantial if layer $l$ is sufficiently far from the output, and the layer pattern has high support.
Notice that if the patterns are formally verified then this hybrid setup is guaranteed to have no degradation in accuracy.
Having said this, we also note that most distillation methods typically tolerate a small degradation in accuracy. Consequently, instead of the expensive formal verification step one could perform an empirical validation of the patterns, and select ones that hold with high probability. This makes the approach practically attractive. As a proof of concept, we evaluate this approach on an eight layer MNIST network in Section~\ref{sec:applications}. Interestingly, we note that a network simplified in this manner satisfies the inferred properties {\em by construction}, without any proof needed.

\section{Computing Network Properties}\label{sec:methods}

We now describe two techniques to build input and layer properties from a feed-forward network wrt convex output property $P$.

\subsection{Iterative relaxation of decision patterns}
This is a technique for extracting input properties.
It makes use of an off-the-shelf decision procedure for neural networks. In this work, we use Reluplex~\cite{KaBaDiJuKo17Reluplex} but other decision procedures can be used too (see Section~\ref{sec:related}).
\footnote{As discussed, in the absence of a decision procedure, empirical validation of properties can also used. While we would lose the formal guarantee that the computed decision patterns imply the postcondition, they may still be useful in practice.}

Recall from Section~\ref{sec:theory} that an input property is a $\prec$-closed pattern $\dpattern$ that satisfies $\forall X: \dpattern(X) \implies P(F(X))$. Ideally we would like to identify the weakest such pattern, i.e., one that constraints the fewest neurons. Computing such a property would involve enumerating all  $\prec$-closed patterns ($O(2^{\vert\neurons\vert})$), and using a decision procedure to validate whether Equation~\ref{eqn:defprop} holds. This is computationally prohibitive.

Instead, we apply a greedy approach to identify a {\em minimal} $\prec$-closed pattern $\dpattern$, meaning that there is no $\prec$-closed sub-pattern of $\dpattern$ that also satisfies Equation~\ref{eqn:defprop}. We start with an input $X$ whose output satisfies the postcondition $P$, i.e., $P(F(X))$ holds.
Let $\dpattern_X$ be the \emph{activation signature} (see Definition~\ref{def:activation_sig}) of the input $X$.
By Proposition~\ref{prop:activation_sig} (Part (B)), we have that
$\dpattern_X(X') \wedge P(F(X'))$ is an input property; recall that $P$ is assumed to be convex. But this property may not be minimal. Therefore, we iteratively drop constraints from it till we obtain a minimal property. The algorithm is formally described in the Appendix (see Algorithm 1). It is easy to see that the resulting pattern is $\prec-closed$, minimal, and it implies the output property ($F(X')=y$).
\ignoreme{
We use the notation $\decproc(A(X), B(X))$ for a call to the decision procedure to check $\forall X: A(X) \implies B(X)$.\footnote{Most decision procedures would validate this property by checking whether the formula $A(X) \wedge \neg B(X)$ is satisfiable.} We use $\dpattern ~\setminus~\neurons$ to denote a sub-pattern of $\dpattern$ wherein all neurons from $\neurons$ are unconstrained.

First, we drop the constraint $(F(X')=y)$ and invoke the decision procedure to check if $\dpattern_X(X') \implies F(X')=y$ (line 2). If this does not hold then $\dpattern_X(X') \wedge (F(X')=y)$ is returned as an input invariant; recall that $P$ is assumed to convex. Otherwise, we have that $\dpattern_X(X')$ is an input invariant. We then proceed to unconstrain neurons in $\dpattern_X$. This is done in two steps: (1) We unconstrain all neurons in a single layer, starting from the last layer, until we discover a critical layer $cl$ such that unconstraining all neurons in $cl$ invalidates the invariant (lines 7,8). (2) Within the critical layer $cl$, we unconstrain neurons one at a time, in some arbitrarily picked order, till we end up with a minimal subset (line 12--17).
}

\begin{proposition}\label{prop:iterative_relaxation}
Algorithm 1 (refer Appendix) always returns a minimal input property, and involves at most $n+m$ calls to the decision procedure, where $n$ is the number of layers, and $m$ is the maximum number of neurons in a layer.
\end{proposition}

\stitle{Example} Consider the example network from Figure~\ref{fig:example}, and the input $X=[1.0,-1.0]$ for which the network predicts class~$1$.
We apply Algorithm 1 to identify an input property for class~$1$. The algorithm starts with the \emph{activation signature} of $X$, which is the pattern $\dpattern_X = \{N_{1,1} \rightarrow \on, N_{1,2} \rightarrow \off, N_{2,1}\rightarrow \on, N_{2,2} \rightarrow \off\}$. Notice that $\dpattern_X$ is already an input property for class $1$. The algorithm begins to unconstrain all neurons in each layer, starting from the last layer, and identifies layer~$1$ as the critical layer (i.e., unconstraining neurons in layer~$1$ violates the postcondition). The algorithm 
then identifies $\{N_{1,1} \rightarrow \on, N_{1,2} \rightarrow \off\}$ as a minimal pattern that implies the postcondition.


\subsection{Mining layer properties using decision tree learning}
\label{subsec:dectree}
The greedy algorithm described in the previous section is computationally expensive as it invokes a decision procedure at each step. We now present a relatively inexpensive technique that relies on data, and avoids invoking a decision procedure  multiple times. The idea is to observe the activation signatures of a large number of inputs, and \emph{learn} decision patterns that imply various output properties.
In this work, we use decision tree learning (see Appendix for background) to extract compact rules based on the activation statuses ($\on$ or $\off$) of neurons in a layer.
Decision trees are attractive as they yield decision patterns that are compact (and therefore have high support) based on various information-theoretic measures.
The resulting patterns are empirically validated layer properties, which can be formally checked with a single call to a decision procedure.

\begin{figure*}[t]
\begin{subfigure}[c]{0.5\textwidth}
\begin{center}
 \begin{xtabular}{|p{0.3\textwidth}|p{0.3\textwidth}|p{0.2\textwidth}|}
\hline
 $\langle x_1,x_2\rangle$ & $\langle N_{1,1},N_{1,2} \rangle$ & $P(F(X))$ \\ [0.5ex]
 \hline
 $\langle 0,-1\rangle$ & $\langle\on,\off\rangle$ & True \\
 \hline
 $\langle1,0\rangle$ & $\langle\on,\on\rangle$ & True \\
 \hline
 $\langle0,1\rangle$ & $\langle\off,\on\rangle$ & False \\
 \hline
 $\langle4,3\rangle$ & $\langle\on,\on\rangle$ & False \\
 \hline
 $\langle1,-1\rangle$ & $\langle\on,\off\rangle$ & True \\
 \hline
 \end{xtabular}
 \caption{Training dataset for decision tree.}\label{fig:dataset}

 \end{center}
\end{subfigure}
\hfill
\begin{subfigure}[c]{0.4\textwidth}
\begin{center}
{\footnotesize

\includegraphics[scale=0.25]{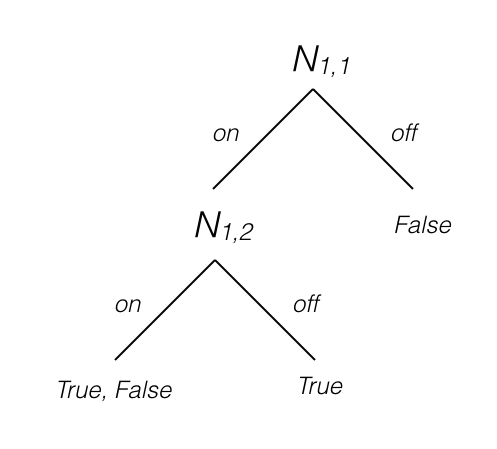}
\vspace{-0.5cm}
\caption{Resultant decision tree. The pattern harvested for True is $\{N_{1,1} \rightarrow \on,  N_{1,2} \rightarrow \off\}$.}\label{fig:tree}
}
\end{center}
\end{subfigure}
\caption{{ Illustration of decision tree learning for mining properties for the network in Figure~\ref{fig:example}. The output property is that the top predicted class is ``1''.}\label{fig:dectreeexampleillustration}}
\end{figure*}

Our algorithm works as follows.  Suppose we have a dataset of inputs $\D$. Consider a layer $l$ where we would like to learn a layer property wrt postcondition $P$. We evaluate the network on each input $X \in \D$, and note: (1) the activation status of all neurons in layer $l$, denoted by $\dpattern_X^l$, and (2) the boolean $P(F(X))$ indicating whether the output $F(X)$ satisfies property $P$. Thus, we have a labeled dataset of feature vectors $\dpattern_X^l$ mapped to labels $P(F(X))$; see for example Figure~\ref{fig:dataset}. We now learn a decision tree from this dataset. The nodes of the tree are neurons from layer $l$, and branches are based on whether the neuron is $\on$ or $\off$. Each path from root to a leaf labeled \textsf{True} forms a decision pattern for predicting the output property; see Figure~\ref{fig:tree}.
We filter out  patterns $\dpattern$ that are \emph{impure}, meaning that there exists an input $X \in \D$ that satisfies $\dpattern(X)$ but $P(F(X))$ does not hold.
The remaining patterns are ``likely'' layer properties wrt the postcondition. We sort them in decreasing order of their support and invoke the decision procedure ($\decproc(\dpattern(X), P(F(X)))$) to formally verify them. This last step can be skipped for applications such as distillation (see Section~\ref{sec:applications}) where empirically validated patterns may suffice.

We can refine the method for the case where the output property is a prediction postcondition i.e., of the form $P(Y) \mydef argmax(Y) = c$. In this case, rather than predicting a boolean as to whether the predicted class is $c$, we train a decision tree to directly predict the class label. This lets us harvest layer patterns for prediction postconditions corresponding to all classes. Specifically, the path from the root to a leaf labeled class $c$ is a likely layer property for the postcondition that the top predicted class is $c$.

\stitle{Counter-example guided refinement}
In verifying Equation~\ref{eqn:defprop} for a decision pattern $\dpattern$ using a decision procedure, if a counter-example is found, we strengthen the pattern by additionally constraining the activation status of those neurons from layer $l$ that have the same activation status for all inputs satisfying the pattern $\dpattern$. If verification fails on this stronger pattern then we do a final step of constraining {\em all} neurons from layer $l$ based on the activation signature of a \emph{single} input satisfying the pattern. If verification still fails, we discard the pattern.
One can also consider a different strategy for refinement, were the counter-examples are added back to the data set and the decision tree learning is re-run, obtaining new layer patterns that will no longer lead to those counter-examples. The drawback is that it may require too many calls to the decision procedure, if many refinement steps are needed.



\section{Applications}\label{sec:applications}
In this section, we discuss case studies on computing input and layer properties, and using them for different applications.
We implemented all our algorithms
in Python 3.0 and Tensorflow.  The Python notebook is connected to  Python2 Google Compute Engine backend with 12Gb RAM allotted.
Our implementation supports analysis of both ReLU and CNN networks. However, for the proofs we use Reluplex~\cite{KaBaDiJuKo17Reluplex}, which is limited to ReLU networks.  To enforce a decision pattern we modified Reluplex to constrain intermediate neuron values. As more decision procedures for neural networks become available, we plan to incorporate them in our tool, thus extending its applicability. The Reluplex runs were done on a server with Ubuntu v16.04 (8 core, 64 GB RAM). We use the linear programming solver {\tt pulp 2.3.1} to solve for under-approximation boxes.
We plan to make the implementation and the networks available with a final paper version.

\subsection{Analysis of ACASXU}
\label{sec:acasxu}

We first discuss the analysis of {\bf ACASXU}, a safety-critical collision avoidance system for unmanned aircraft control~\cite{ACASXU}.
ACASX is a family of collision avoidance systems for aircraft, under development by the Federal Aviation Administration (FAA). ACASXU is the version for unmanned aircraft. It receives sensor information regarding the drone (the \emph{ownship}) and any nearby intruder drones, and then issues horizontal turning advisories aimed at preventing collisions.
The input sensor data includes:
(1) Range: distance between ownship and intruder;
(2) $\theta$: angle of intruder relative to ownship heading direction;
(3) $\psi$: heading angle of intruder relative to ownship heading direction;
(4) $v_{\text{own}}$: speed of ownship;
(5) $v_{\text{int}}$: speed of intruder;
(6) $\tau$: time until loss of vertical separation; and
(7) $a_{\text{prev}}$: previous advisory.

The FAA is exploring an implementation of ACASXU that uses an array of 45 deep neural networks, from which we selected one network for discussion here.
The five possible output actions are as follows: (0) Clear-of-Conflict (COC), (1) Weak Left, (2) Weak Right, (3) Strong Left, and (4) Strong Right. Each advisory is assigned a score, with the lowest score corresponding to the best action.
The network that we analyzed (namely model 1\_1) consists of 6 hidden layers, and 50 ReLU activation nodes per layer.
We used 384221 inputs with known labels. ACASXU networks were analyzed before with Reluplex \cite{KaBaDiJuKo17Reluplex}. Verification for ACASXU is challenging, taking many hours (may even time  out after 12h); we give more details below.

%
%
%

\subsubsection{Property Inference}
We infer network properties wrt prediction postconditions that require that the output of a network classifier is a certain class.
We used decision tree learning to extract layer patterns; we list them all (total 25) in Table III in Appendix. The learning took 45 seconds on average per layer (4.5 minutes in total). We discuss here the verification of one specific layer pattern. This pattern was for label COC (clear-of-conflict) at layer $5$, and was subsequently used to decompose proofs of ACASXU properties, as discussed below. The pattern has a support of 109417 inputs. We were able to prove a property computed based on this pattern after two refinement steps (Section~\ref{sec:methods}), within 5 minutes. We also extracted candidate input properties corresponding to the decision pattern of the layer property following proposition~\ref{prop:layerinvariant}. From the 109417 inputs that satisfied the decision pattern at layer 5, we extracted distinct decision prefixes corresponding to 5532 inputs. We were able to prove all of them (3600 properties) within an average time of 1 minute per property.

These experiments show that it is feasible to extract input and layer properties in terms of the on/off patterns of the ReLU nodes of real networks.
The experiments also show that the patterns constraining lesser number of neurons have higher support and layer properties have higher support than input properties, as expected, since they cover a union of regions in the input space.

\subsubsection{Explaining Network Predictions}
The input-output properties derived for ACASXU can explain the network behavior.  We further used LP solving to calculate  under-approximation boxes corresponding to  input properties. We calculated such a box for each of the 3600 input properties that we had proved.
We also generated under-approximation boxes for input decision patterns that could not be proved within a time limit of 12 hours but had high support. This helped elicit novel properties of the network, which were validated by the domain experts. We give some examples below.

-- All the inputs within: \textit{31900 $\leq$ range $\leq$ 37976,  1.684 $\leq$ $\theta$ $\leq$ 2.5133, $\psi$ $=$ -2.83, 414.3 $\leq$ $v_{\text{own}}$ $\leq$ 506.86, $v_{\text{int}}$ $=$ 300}, should have the turning advisory as COC.

-- All the inputs within: \textit{range $=$ 499,  -0.314 $\leq$ $\theta$ $\leq$ -3.14, -3.14 $\leq$ $\psi$ $\leq$ 0, 100 $\leq$ $v_{\text{own}}$ $\leq$ 571, 0 $\leq$ $v_{\text{int}}$ $\leq$ 150}, should have the turning advisory as Strong Left.


We further experimented with computing minimal assignments that satisfy the inferred properies.
For instance, we analyzed a layer $2$ property for the label COC, with a support of 51704 inputs.
By computing the minimal assignment over an input that satisfied this property, we determined that the last two input attributes, namely, $v_{\text{own}}$ (speed of ownship) and $v_{\text{int}}$ (speed of intruder) were not relevant when the other attributes are constrained as follows: range $=$ 48608,  $\theta$ $=$ -3.14 and $\psi$ $=$ -2.83. This represents an input-output property of the network elicited by our technique. The domain experts confirmed that this was indeed a valid and novel property of the ACASXU network. 

\subsubsection{Layer Patterns as Interpolants}
To evaluate the use of layer patterns in simplifying difficult proofs, we selected 3 properties from the ACASXU application. These properties have previously been considered for verification directly using Reluplex ~\cite{KaBaDiJuKo17Reluplex}.
We list here the three properties.
\begin{itemize}
    \item{Property 1:}  All inputs within the following region, \textit{36000 $\leq$ range $\leq$ 60760, 0.7 $\leq$ $\theta$ $\leq$ 3.14, -3.14 $\leq$ $\psi$ $\leq$ -3.14 + 0.01, 900 $\leq$ $v_{\text{own}}$ $\leq$ 1200, 600 $\leq$ $v_{\text{int}}$ $\leq$ 1200}, should have the turning advisory as COC. This property takes approx. 31 minutes to check with Reluplex. 
    \item{Property 2:} All the inputs within the following region: \textit{12000 $\leq$ range $\leq$ 62000, (0.7 $\leq$ $\theta$ $\leq$ 3.14) or (-3.14 $\leq$ $\theta$ $\leq$ -0.7), -3.14 $\leq$ $\psi$ $\leq$ -3.14 + 0.005,   100 $\leq$ $v_{\text{own}}$ $\leq$ 1200,  0 $\leq$
 $v_{\text{int}}$ $\leq$ 1200}, should have the turning advisory as COC. This property has a large input region and direct verification with Reluplex times out after 12 hours.
    \item{Property 3:} All the inputs within the following region: \textit{range $>$ 55947.691, -3.14 $\leq$ $\theta$ $\leq$ 3.14, -3.14 $\leq$ $\psi$ $\leq$ 3.14, 1145 $\leq$ $v_{\text{own}}$ $\leq$ 1200, 0 $\leq$ $v_{\text{int}}$ $\leq$ 60}, should have the turning advisory
 as Clear-of-Conflict (COC). This property takes approx. 5 hours to check with Reluplex.
\end{itemize}

All three properties have the form $A\implies B$, where $A$ specifies constraints on the input attributes, and $B$ specifies that the output turning advisory is COC. For each property, we used decision tree learning to extract multiple layer patterns for label COC at every layer, and selected the one that covers maximum number of inputs within the input region $A$. Incidentally, for all three properties the same pattern at layer $5$ (denoted by $\dpattern^5$) was selected.

\paragraph{Property 1:} We found 195 inputs in the training set that fall within $A$ and classify as COC. All of these inputs are also covered by $\dpattern^5$.  We therefore proceeded to prove $A \implies \dpattern^5$ and $\dpattern^5 \implies B$ using Reluplex. For proving $\dpattern^5 \implies B$, we had to strengthen the pattern by constraining 48 nodes at layer 5. This made the proof go through, and finish in 5 minutes.

We then attempted to prove $A \implies \dpattern^5$ for the strengthened version. This process finished in 2 minutes. Thus, we were able to prove this property in 7 minutes. In contrast, direct verification of the property using Reluplex takes 31 minutes.

\paragraph{Properties 2 and 3:} We could not identify a single layer pattern that covered the inputs within $A$ completely. The pattern $\dpattern^5$ had maximum coverage with respect to the training inputs within $A$ ($5276/7618$ inputs for property 2, $256/441$ for property 3).  We split the proof into two parts. First, we extracted the activation signature prefixes up to layer $5$ for each of the training inputs that satisfy $\dpattern^5$.
Let $\mi{cov}$ be the set of these prefixes.
We then checked $(A \wedge  \bigvee_{\dpattern_i \in \mi{cov}}\dpattern_i(X)) \implies B$\footnote{Since $\dpattern^5$ implies the property $B$ only after strengthening, showing that $(A \wedge  \bigvee_{i \in \mi{cov}}\dpattern_i(X)) \implies \dpattern^5$ is not enough to ensure that $(A \wedge  \bigvee_{i \in \mi{cov}}\dpattern_i(X)) \implies B$.}
Checks of the form $(A \wedge  \dpattern_i(X)) \implies B$ were spawned in parallel  for every $\dpattern_i$. This completed in an hour for property 2 and within 6 mins for property 3.
The remaining obligation in completing the proof for the property was $(A \wedge \neg(\bigvee_{i \in cov}  \dpattern_i(X))) \implies B$.
To check this efficiently, we determined the under-approximation boxes for each $\dpattern_i$, and spawned parallel checks on the partitions within $A$ not covered by the boxes.
The longest time taken by any job was 2 hours 10 minutes for property 2 and 1 hour 30 minutes for property 3.
 This is a promising result as a direct proof of property 2 using Reluplex times out after 12 hours. For property 3, a direct proof takes 5 hours.


\subsection{Analysis of MNIST}

We also analyzed {\bf MNIST}, an image classification network based on a large collection of handwritten digits~\cite{MNISTWebPage}. 
It has 60,000 training input images, each characterized by 784 attributes and belonging to one of 10 labels. We first analyzed a simple network from the Reluplex distribution (containing 10 layers with 10 ReLU nodes per layer). The simplicity of the network makes it amenable to proofs using Reluplex. For the distillation experiments (described in the following subsection) we use a more complex MNIST network that is close to state-of-the art.

\subsubsection{Property Inference} We extracted input properties using iterative relaxation and layer properties using decision tree learning, showing the feasibility of our approach in the context of image classification, which involves a much larger input space compared to ACASXU. Details about the computed properties (total 30) are given in Tables I and II in the Appendix.

\begin{figure}[t]
\begin{center}
\includegraphics[scale=0.7]{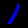}
\includegraphics[scale=0.7]{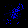}


\includegraphics[scale=0.7]{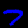}
\includegraphics[scale=0.7]{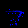}


\includegraphics[scale=0.7]{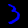}
\includegraphics[scale=0.7]{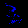}

\caption{{Original images from the data set (left). Counter-examples to failed proofs for patterns containing the original images (right).} \label{fig:ADV_1}}
\end{center}
\end{figure}

The Reluplex checks for some of the network properties generated counter-examples which show potential vulnerabilities of the network, since they are close (in the Euclidean space) to other inputs that are classified differently (Figure \ref{fig:ADV_1}).

\begin{figure}[t]
\begin{center}
\includegraphics{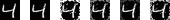}
\includegraphics{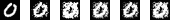}
\includegraphics{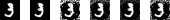}
\end{center}
\caption {Visualization of MNIST input properties using under-approximation boxes.
\label{fig:MNISTInputInvars}}
\end{figure}

\subsubsection{Explaining Network Predictions} We further computed and visualized under-approximation boxes for the inferred properties. As an example, in Figure~\ref{fig:MNISTInputInvars}, we show a visualization of input properties corresponding to three different images from the training set. The first column shows original images.
Columns 2 and 3  show images with all pixels set to their minimum and maximum values in the computed under-approximating box, respectively.
Columns 4, 5 and 6 have each pixel set to the mean value of its range in the box, a randomly chosen value below the mean, and a randomly chosen value above the mean, respectively.

\begin{figure}[t]
\begin{center}
\includegraphics[scale=0.7]{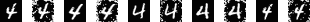}
\includegraphics[scale=0.7]{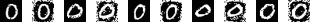}
\includegraphics[scale=0.7]{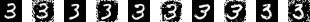}
\ignoreme{\includegraphics{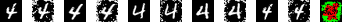}
\includegraphics{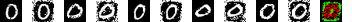}
\includegraphics{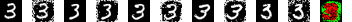}}
\end{center}
\caption {Visualization of MNIST layer properties using under-approximation boxes.
\label{fig:MNISTInterInvars}}
\end{figure}

In Figure~\ref{fig:MNISTInterInvars}, we visualize layer properties via under-approximation boxes corresponding to 5 input properties, based on 5 randomly chosen images from the support of the property. Each box is represented by 2 images, setting all the pixels to their respective minimum and maximum values in the box.
Note that the images drawn from the under-approximation boxes represent {\em new inputs} (not in the training set) that satisfy the same property and hence are labelled the same. While input properties capture visually (or syntactically) similar images, layer properties cluster images of the same digit written in different ways, indicating that layer properties can potentially capture common features across inputs.
The developer can examine the generated images to get a sense of the image characteristics that contributed to the network decisions.

\begin{figure}[t]
\begin{minipage}[c]{0.4\textwidth}
\begin{center}
\includegraphics[scale=0.8]{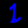}
\includegraphics[scale=0.8]{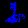}
\includegraphics[scale=0.8]{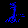}
\caption {{Digit 1 misclassified to 2 and images with min and max values from under-approximation box of original image.} \label{fig:MIS1_1}}
\end{center}
\end{minipage}
\end{figure}
\subsubsection{Misclassifications} Under-approximation boxes can also be used to reason about {\em misclassifications}. Misclassified inputs are typically ``rare" and spread across the input space, and it is very difficult for developers to understand their cause and fix the underlying problem.  Figure~\ref{fig:MIS1_1} shows an image of digit 1 misclassified to digit 2 (Figure~\ref{fig:MIS1_1}, first column). We used this input to extract an input decision pattern and compute an under-approximation box for it (Figure~\ref{fig:MIS1_1}, 2nd and 3rd columns). We can thus draw many more inputs from the box that are similarly misclassified. These inputs can help developers understand the cause of misclassification and re-train the network on them.


\subsection{Distillation}
\ignoreme{
\begin{figure}[t]
\begin{minipage}[c]{0.50\textwidth}
\includegraphics[scale=0.25]{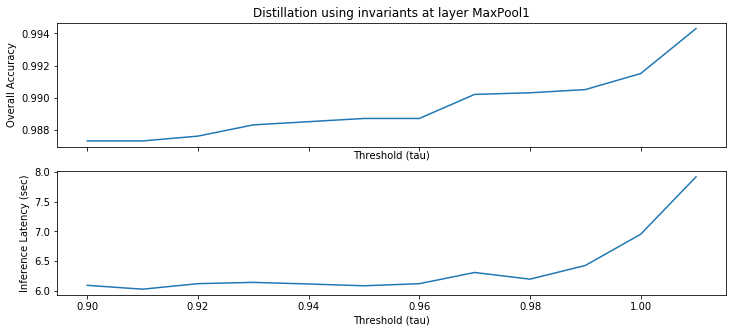}
\end{minipage}
\hfill
\begin{minipage}[c]{0.50\textwidth}
\includegraphics[scale=0.25]{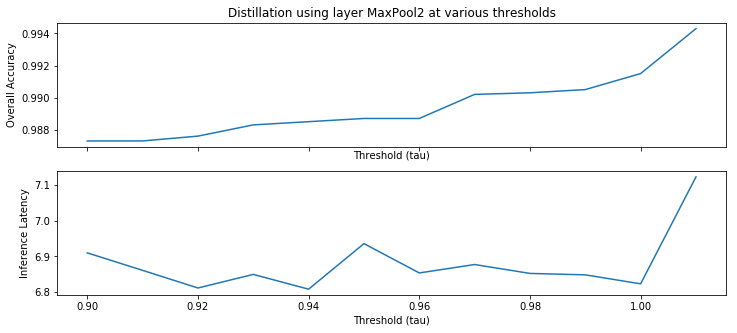}
\end{minipage}
\caption{Distillation of an eight layer MNIST network (from~\cite{Carlini017}) using layer properties from two different max pooling layers. The x-axis shows the empirical validation threshold used for selecting properties. The extreme right point (threshold $>$ 1) corresponds to one where no properties are selected, and therefore distillation is not triggered. The reported inference times are based on an average of 10 runs of the test dataset on a single core Intel(R) Xeon(R) CPU @ 2.30GHz.}\label{fig:distillation}
\end{figure}
}
\begin{figure}[t]
\begin{center}
\includegraphics[scale=0.35]{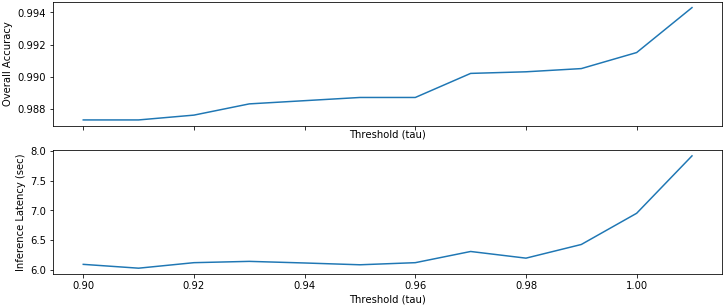}
\end{center}
\caption{Distillation of an eight layer MNIST network (from~\cite{Carlini017}) using properties at the first max pooling layer. }\label{fig:distillation1}
\end{figure}

\begin{figure}[t]
\begin{center}
\includegraphics[scale=0.35]{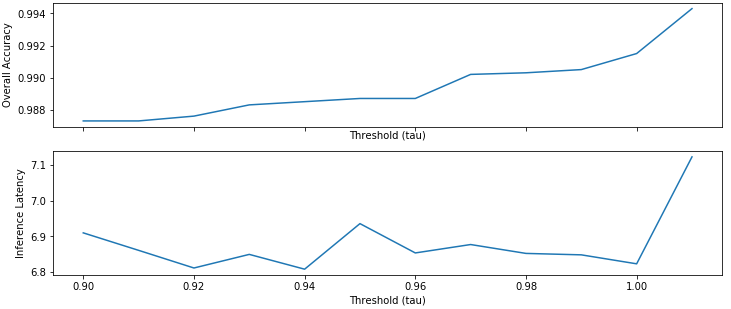}
\end{center}
\caption{Distillation of an eight layer MNIST network (from~\cite{Carlini017}) using layer patterns at the second max pooling layers.}\label{fig:distillation2}
\end{figure}

Our final experiment is to evaluate the use of layer properties in distilling a network. As discussed in Section~\ref{subsec:interpreting_invariants}, the key idea is to use prediction properties at an intermediate layer as distillation rules. For inputs satisfying the property, we save the inference cost of evaluating the network from the intermediate layer onwards. We present a preliminary evaluation of this idea using a more complex MNIST network~\cite{Carlini017} with 8 hidden layers; two convolutional, one max pooling, two convolutional, one max pooling, and  two fully connected layers. The network has a superior accuracy of $0.9943$ but it is computationally expensive during inference. We use the decision tree algorithm to obtain layer patterns. We then empirically validate them (using a validation set of $5000$ images), and select ones with accuracy above a threshold $\tau$ (see Section~\ref{sec:theory}). The selected properties are used as distillation rules for inputs satisfying them. Using a held-out test dataset, we measure the overall accuracy and inference time of this hybrid setup for different values of $\tau$.

Figure~\ref{fig:distillation1} shows the results of distillation from the first max pooling layer\footnote{While max pooling neurons are different from ReLU neurons, we could still consider activation patterns on them based on whether the neuron output is greater than $0$ or equal to $0$. A decision tree can then be learned over these patterns to fit the prediction labels.}, which consists of $4608$ neurons.
The x-axis shows the empirical validation threshold used for selecting properties. The extreme right point (threshold $>$ 1) corresponds to one where no properties are selected, and therefore distillation is not triggered. The reported inference times are based on an average of 10 runs of the test dataset on a single core Intel(R) Xeon(R) CPU @ 2.30GHz.
The figure shows the trend of overall accuracy and inference time as the threshold $\tau$ is varied from $0.9$ to $1.0$. Observe that at a threshold of $\tau = 0.98$, one can achieve a 22\% saving in inference time while only degrading accuracy from $0.9943$ to $0.9903$. This is quite promising. As expected, lowering the threshold further considers more properties, and therefore reduces both inference time and accuracy. The results from the second max pooling layer (shown in Figure~\ref{fig:distillation2}) are similar except that both the degradation in accuracy and the saving in inference time are smaller. This is expected as the second max pooling layer is closer to the output, and therefore the properties that we infer approximate a smaller part of the network.

\section{Related Work}\label{sec:related}
We survey the works that are the most closely related to ours.
In~\cite{SzegedyZSBEGF13} it has been shown that neural networks are highly vulnerable
to small adversarial perturbations. Since then, many works have focused on methods for
finding adversarial examples. They range from heuristic and optimization-based methods~\cite{GoodfellowSS14,KurakinGB16a,PapernotMJFCS16,Carlini017,Moosavi-Dezfooli16} to analysis techniques which can also provide
formal guarantees. In the latter category, tools based on constraint solving, interval analysis or abstract interpretation, such as DLV~\cite{HuangKWW17}, Reluplex~\cite{KaBaDiJuKo17Reluplex}, AI$^2$\cite{DBLP:conf/sp/GehrMDTCV18}
ReluVal~\cite{WangPWYJ18F}, Neurify~\cite{WangPWYJ18E} and others~\cite{DuttaJST18,abs-1712-06174}, are gaining prominence. Our work is complementary as it focuses on inferring input-output properties of neural networks. In principle, we can leverage the previous analysis techniques to verify the inferred properties.

There are several papers on explaining predictions made by neural networks, see~\cite{GMRTGP18} for a survey. One line of work is on explaining individual predictions by attributing them to input features~\cite{SimonyanVZ13,RSG16,ShrikumarGSK16,SundararajanTY17,LL17,KSAMEKD18}.
They are either based on computing gradients of the prediction with respect to input features~\cite{SimonyanVZ13,SundararajanTY17}, back-propagating the prediction score to input features using a set of rules~\cite{ShrikumarGSK16,KSAMEKD18}, using attribution techniques from cooperative game theory~\cite{LL17}, or computing local linear approximations of the behavior of the network~\cite{RSG16}.

The closest to ours is the work on Anchors~\cite{Ribeiro0G18}, which aims to explain the network behaviour by means of {\em rules} (called {\em anchors}), which represent sufficient conditions for network predictions. These anchors are computed solely based on the black-box behaviour of the neural network. Input properties from our work can be viewed as anchors for various output properties. The key difference is that our input properties are obtained via a white-box analysis of the neurons in the network, and are backed with a formal guarantee.

Also relevant, there is work on computing the influence of individual neurons on predictions made by the network~\cite{DSY19,LSDFL18}. In a sense, our layer properties can be seen as a means for identifying influential neurons for a prediction, the key difference being that layer properties also guarantee that decisions of the influential neurons indeed imply the prediction. These previous approaches evaluate neuron influence by measuring how accurately the top k most influential neurons alone can predict the class. Interestingly, we believe these works also lend themselves to distillation.
We leave a thorough comparison of different distillation mechanisms to future work.

There is a large body of work on property inference, including ~\cite{Flanagan:2001:HAA:647540.730008,ErnstPGMPTX2007,DBLP:conf/cav/0001LMN14,Garg:2016:LIU:2837614.2837664} to name just a few, although none of the previous works have addressed neural networks.
The programs considered in this literature tend to be small but have complex constructs such as loops, arrays, pointers. In contrast, neural networks have simpler structure but can be massive in scale.

A recent paper~\cite{NIC} uses properties over neuron activation distributions to determine whether a given input is benign (i.e., non adversarial).
The `invariants' in ~\cite{NIC} are meant to capture properties of a given set of inputs (benign inputs), while our input and layer properties are meant to capture properties of the network. Furthermore, our properties partition the input space into prediction-based regions, and are justified with a formal proof.
We do note that our properties can be seen as {\em invariance} properties of the network, that have the special form `precondition implies postcondition'.

Our distillation approach is related to teacher-student learning in neural networks~\cite{teacher-student}. Note that we do not perform transfer learning (from a teacher to a student) but instead use the inferred properties to simplify the network. Thus, unlike teacher/student learning, our distillation approach is {\em adaptive}, allowing to process some inputs (that satisfy the layer properties) using the simplified computation; the other inputs (that may need more complex processing) go through the original network. Furthermore, we provide formal guarantees as by construction, our `distilled' network satisfies the properties used in the distillation.

\section{Conclusion}
We presented techniques to extract neural network input-output properties
and we discussed their application to explaining neural networks,
providing robustness guarantees,
simplifying proofs and distilling the networks.
As more decision procedures for neural networks become available, we plan to incorporate them in our tool, thus extending its applicability and scalability.
We also plan to leverage the decision patterns to obtain parallel verification techniques for neural networks and to investigate other applications of the inferred properties, such as confidence modeling, adversarial detection and guarding monitors for safety and security critical systems.

\bibliographystyle{IEEEtranS}
\bibliography{bib,bib2}

\section{Appendix:}

\subsection{Proof for Theorem 1}

\begin{theorem}\label{thm:convexity}
For all $\prec$-closed patterns $\dpattern$, $\dpattern(X)$ is convex, and has the form:
\begin{equation*}
\bigwedge_{i \, in \, 1 .. \vert\on(\dpattern)\vert} W_i\cdot X + b_i > 0 ~ \wedge ~ \bigwedge_{j \, in \, 1 .. \vert\off(\dpattern)\vert} W_j\cdot X + b_j \leq 0
\end{equation*}
Here $W_i, b_i, W_j, b_j$ are some constants derived from the weight and bias parameters of the network.
\end{theorem}

We prove the following stronger property: For all neurons $N$ in a $\prec$-closed pattern $\dpattern$, there exist parameters $W, b$ such that:
\begin{equation}\label{eqn:goal}
\forall X: \dpattern(X) \implies N(X) = \relu(W\cdot X + b)
\end{equation}
The theorem can be proven from this property by applying the definition of $\relu$. We prove this property for all neurons in $\dpattern$ by induction over the depth of the neurons. The base case of neurons in layer $1$ follows from the definition of feed forward ReLU networks.
For the inductive case, consider a neuron $N$ in $\dpattern$ at depth $k$.  Let $N_1, \ldots, N_p$ be the neurons that directly feed into $N$ from the layer below. By recursively expanding $N(X)$, we have that there exist parameters $b, w_1, \ldots, w_p$ such that:
\begin{equation}\label{eqn:nrecursive}
N(X) = \relu(w_1\cdot N_1(X) + \ldots + w_p\cdot N_p(X) + b)
\end{equation}
By induction hypothesis, we have that for each $N_i$ (where $1 \leq i \leq p)$, there exists $W_i, b_i$ such that:
\begin{equation}\label{eqn:hypothesis}
\forall X: \dpattern(X) \implies N_i(X) = \relu(W_i\cdot X + b_i)
\end{equation}
Since $\dpattern$ is $\prec$-closed, $N_1, \ldots, N_p$ must be present in $\dpattern$. Without loss of generality, let $N_1, \ldots, N_k$ be marked $\off$, and $N_{k+1}, \ldots, N_p$ be marked $\on$. The definition of $\dpattern(X)$ is as follows,

\begin{equation}\label{eqn:dpattern}
\dpattern(X) \mydef \bigwedge_{N \in \on(\dpattern)} N(X) > 0 ~ \wedge ~ \bigwedge_{N \in \off(\dpattern)} N(X) = 0
\end{equation}

Hence we have:
\begin{equation}\label{eqn:dpatternoff}
\forall i \in \{1, \ldots, k\}~ \forall X: \dpattern(X) \implies N_i(X) = 0
\end{equation}
\begin{equation}\label{eqn:dpatternon1}
\forall i \in \{k+1, \ldots, p\}~ \forall X: \dpattern(X) \implies N_i(X) > 0
\end{equation}
From Equations~\ref{eqn:dpatternon1} and \ref{eqn:hypothesis}, and definition of $\relu$, we have:
\begin{equation}\label{eqn:dpatternon2}
\forall i \in \{k+1, \ldots, p\} ~\forall X: \dpattern(X) \implies N_i(X) = W_i\cdot X + b_i
\end{equation}
Using Equations \ref{eqn:nrecursive}, \ref{eqn:dpatternoff}, and \ref{eqn:dpatternon2}, we can show that there exists  parameters $W$ and $b$ such that
\begin{equation}\label{eqn:goal}
\forall X: \dpattern(X) \implies N(X) = \relu(W\cdot X + b)
\end{equation}
This proves the property for neuron $N$. 

\subsection{Background on Decision Tree Learning}
Decision tree learning is a supervised learning technique for extracting rules that act as classifiers. Give a set of data \{$D$\} respective classes \{$c$\}, decision tree learning aims to discover rules ($R$) in terms of the attributes of the data to discriminate one label from the other. It builds a tree such that, each path of the tree is a rule $r$, which is a conjunction of predicates on the data attributes. Each rule attempts to cluster or group inputs that belong to a certain label.
There could be more than one paths leading to the same label, and therefore more than one rules for the same label. The decision tree learning attempts to extract compact rules that are generalizable beyond the training data. Each rule is {\em discriminatory} within the tolerance of the error threshold (i.e. two inputs with different labels will have different rules).

$$Tree = DecTree(\{data\},\{label\}, \{attribute\})$$
$$Tree \mydef ( path_{0} \lor path_{1} \lor ... \lor path_{n} )$$
$$R \mydef \{ r_{0} , r_{1} , ... , r_{n} \}$$
$$r_{i} = \bigwedge_{a \in \{attribute\}} predicate(a)$$

\subsection{Algorithms and Results}
We present the iterative relaxation of decision patterns in Algorithm~\ref{alg:itrrel}.
We use the notation $\decproc(A(X), B(X))$ for a call to the decision procedure to check $\forall X: A(X) \implies B(X)$.\footnote{Most decision procedures would validate this property by checking whether the formula $A(X) \wedge \neg B(X)$ is satisfiable.} We use $\dpattern ~\setminus~\neurons$ to denote a sub-pattern of $\dpattern$ wherein all neurons from $\neurons$ are unconstrained.

First, we drop the constraint $(F(X)=y)$ and invoke the decision procedure to check if $\dpattern(X) \implies P(F(X))$. If this does not hold (i.e. $\dpattern(X) \wedge \neg P(F(X))$ is satisfiable) then $\dpattern(X) \wedge P(F(X))$ is returned as an input property. Otherwise, we have that $\dpattern(X)$ is an input property. However it may be too `narrow'. We then proceed to unconstrain neurons in $\dpattern$ while still maintaining the output property. This is done in two steps: (1) We unconstrain all neurons in a single layer, starting from the last layer, until we discover a critical layer $cl$ such that unconstraining all neurons in $cl$ no longer implies the postcondition. (2) Within the critical layer $cl$, we unconstrain neurons one by one (the exact order does not matter as all the conditions in a layer are independent) till we end up with a minimal subset.
\begin{algorithm}
	\begin{algorithmic}[1]
	    \State // Let k be the layer before output layer
	    \State // We write $\neurons^l$ for the neurons at layer $l$
	    \State $\dpattern = \dpattern_X$ // Activation signature of input $X$
	    \State $sat = {\bf DP} (\dpattern(X),P(F(X)))$
	    \If{$sat$}
	    \Return {$\dpattern(X) \wedge P(F(X))$}
	    \EndIf
	    \State $l=k$
	    \While{$l > 1$}
	    \State $\dpattern = \dpattern~\setminus~\neurons^l$
	    \State $sat = {\bf DP} (\dpattern(X), P(F(X))$
	    \If{$sat$}
	    \State //~Critical layer found
	    \State $cl=l$
	    \State // Add back activations from critical layer
	    \State $\dpattern = \dpattern~\cup~\neurons^{cl}$
	    \ForEach{$N \in \neurons^{cl}$}
	    \State $\dpattern' = \dpattern~\setminus~\{N\}$
	    \State $sat = {\bf DP} (\dpattern'(X), P(F(X))$
	    \If{$\neg sat$}
	    \State // Neuron $N$ can remain unconstrained
	    \State $\dpattern = \dpattern'$
	    \EndIf
	    \EndFor
	    \State \Return{$\dpattern(X)$}
	    \Else
	    \State $l=l-1$
	    \EndIf
	    \EndWhile
	\end{algorithmic}
	\caption{Iterative relaxation algorithm to extract input properties from input $X$. \label{alg:itrrel}}
\end{algorithm}

We now describe how we use an LP solver to compute an under-approximation box containing inputs satisfying an input property.
Given an input property, $\dpattern$, we aim to compute a box represented as ranges on each input attribute ($0 ... n$), such that every point within this box which is an input to the network which would satisfy  $\dpattern$; in other words the box is an under-approximation of the property. The code snippet below shows the use of the pulp LP solver to determine these ranges.

\textit{\begin{algorithmic}
\For{i=0; i$<$n;i++}
	    \State $d_{hi}$=pulp.LpVar(name,lowBound=min[i], \newline upBound=max[i],cat=Continuous)
	    \State $d_{lo}$=pulp.LpVar(name,lowBound=min[i],
	    \newline upBound=max[i],cat=Continuous)
	    \State pulpInput[i]=$(d_{lo},d_{hi})$
\EndFor
\end{algorithmic}}

Each input  is represented as a pair of pulp variables, $(d_{lo},d_{hi})$. The variable $d_{lo}$ represents the lowest value and $d_{hi}$ represents the highest value for a given input in the box.  The lower and upper bounds for $d_{lo}$ and $d_{hi}$ are set based on the minimum and maximum values for that attribute in the set of inputs in the support of $\dpattern$. The solver is invoked to maximize the difference between $d_{hi}$ and $d_{lo}$, in order to obtain the widest region possible.

\textit{\noindent{prob=pulp.LpProblem("Box",pulp.LpMaximize)}}

\textit{\noindent{prob+=pulp.lpSum([(pulpInput[i][1]-pulpInput[i][0])],}}

\textit{for[i in range (0,n)]}

Each neuron activation in $\dpattern$ represents a linear constraint imposed by the ReLU activation function on the input variables. For example, at the first hidden layer if for a ReLU node, $N_{0}$ $=$ 0 or $N_{0}$ in $\off(\dpattern)$, indicates that $(\sum_{i=0}^{i=n} x_{i} . w_{i}) + b \leq 0$, where $x_{i}$ is an input variable, $w_{i}$ is the weight of the incoming edge and $b$ is the bias term. This is added as a path constraint for the LP solver as follows. The constraint is re-arranged into the following inequality $(\sum_{i=0}^{i=n} x_{i} . w_{i}) \leq -b$ ($-(\sum_{i=0}^{i=n} x_{i} . w_{i}) \leq b$, if $N_{0}$ is in $\on(\dpattern)$) . If the co-efficient term of an input variable is negative, the variable is replaced by the respective lower bound ($d_{lo}$), while if the co-efficient term is positive, the variable is replaced by ($d_{hi}$).
The equations for the neurons in the layers below can be similarly expressed in terms of $d_{lo}, d_{hi}$ of the input variables (denoted $F(d_{lo},d_{hi})$). These constraints are added to the problem as shown in the snippet below and finally the solver is invoked.

\textit{\begin{algorithmic}
    \If{$N \in \on(\dpattern)$}
    \State prob += ( -(pulp.lpSum( $F(d_{lo},d_{hi})$ )) $\leq$ bias )
    \EndIf
    \If{$N \in \off(\dpattern)$}
    \State prob += ( (pulp.lpSum( $F(d_{lo},d_{hi})$ )) $\leq$ -bias )
    \EndIf
    \State prob.solve()
\end{algorithmic}}

\subsection{Under-approximation boxes}
We list here some of the under-approximation boxes calculated for the 3600 input decision patterns that were proved on the ACASXU network. In each property, all inputs within the given regions should have the turning advisory as COC.

\begin{enumerate}
    \item \textit{39688 $\leq$ range $\leq$ 60760,  1.57 $\leq$ $\theta$ $\leq$ 1.59, $\psi$ $=$ -3.14, 491 $\leq$ $v_{\text{own}}$ $\leq$ 1176, 242 $\leq$ $v_{\text{int}}$ $\leq$ 382}.
    \item \textit{37087 $\leq$ range $\leq$ 60760, 1.28 $\leq$ $\theta$ $\leq$ 1.57, $\psi$ $=$ -3.14, 837 $\leq$ $v_{\text{own}}$ $\leq$ 1200, 232 $\leq$ $v_{\text{int}}$ $\leq$ 611}.
    \item \textit{38060 $\leq$ range $\leq$ 56778,  1.48 $\leq$ $\theta$ $\leq$ 1.57, $\psi$ $=$ -3.14, 728 $\leq$ $v_{\text{own}}$ $\leq$ 1200, 429 $\leq$ $v_{\text{int}}$ $\leq$ 640}.
    \item \textit{38060 $\leq$ range $\leq$ 60760,  1.40 $\leq$ $\theta$ $\leq$ 1.57, $\psi$ $=$ -3.14, 728 $\leq$ $v_{\text{own}}$ $\leq$ 1200, 553 $\leq$ $v_{\text{int}}$ $\leq$ 640}.
    \item \textit{20134 $\leq$ range $\leq$ 51676,  -3.13 $\leq$ $\theta$ $\leq$ 1.14, $\psi$ $=$ -3.14, 911 $\leq$ $v_{\text{own}}$ $\leq$ 1200, 110 $\leq$ $v_{\text{int}}$ $\leq$ 150}.
    \item \textit{29938 $\leq$ range $\leq$ 60760,  2.61 $\leq$ $\theta$ $\leq$ 2.83, $\psi$ $=$ -3.14, 889 $\leq$ $v_{\text{own}}$ $\leq$ 1200, $v_{\text{int}}$ $=$ 0}.
    \item \textit{18821 $\leq$ range $\leq$ 28766,  2.82 $\leq$ $\theta$ $\leq$ 3.09, $\psi$ $=$ -3.14, 407.5 $\leq$ $v_{\text{own}}$ $\leq$ 943.7, 205.3 $\leq$ $v_{\text{int}}$ $\leq$ 300}.
    \item \textit{35240 $\leq$ range $\leq$ 60760,  $\theta$ $=$ 1.57, $\psi$ $=$ -3.14, 965 $\leq$ $v_{\text{own}}$ $\leq$ 1200, 600 $\leq$ $v_{\text{int}}$ $\leq$ 763}.
    \item \textit{34271 $\leq$ range $\leq$ 52747,  1.18 $\leq$ $\theta$ $\leq$ 1.57, $\psi$ $=$ -3.14, 839 $\leq$ $v_{\text{own}}$ $\leq$ 1200, 600 $\leq$ $v_{\text{int}}$ $\leq$ 680}.
    \item \textit{38439 $\leq$ range $\leq$ 53629,  2.46 $\leq$ $\theta$ $\leq$ 2.83, $\psi$ $=$ -3.14, 217 $\leq$ $v_{\text{own}}$ $\leq$ 434, 900 $\leq$ $v_{\text{int}}$ $\leq$ 1200}.
    \item \textit{40489 $\leq$ range $\leq$ 60760,  1.32 $\leq$ $\theta$ $\leq$ 1.59, $\psi$ $=$ -3.14, 826 $\leq$ $v_{\text{own}}$ $\leq$ 1200, 112 $\leq$ $v_{\text{int}}$ $\leq$ 156}.
    \item \textit{31717 $\leq$ range $\leq$ 60760,  1.25 $\leq$ $\theta$ $\leq$ 1.50, $\psi$ $=$ -3.14, 962 $\leq$ $v_{\text{own}}$ $\leq$ 1171, 999 $\leq$ $v_{\text{int}}$ $\leq$ 1050}.
    \item \textit{44813 $\leq$ range $\leq$ 60760,  1.65 $\leq$ $\theta$ $\leq$ 1.88, $\psi$ $=$ -3.14, 728 $\leq$ $v_{\text{own}}$ $\leq$ 1000, 450 $\leq$ $v_{\text{int}}$ $\leq$ 692}.
    \item \textit{13670 $\leq$ range $\leq$ 25541,  1.87 $\leq$ $\theta$ $\leq$ 2.19, -3.14 $\leq$ $\psi$ $\leq$ -3.13, 571 $\leq$ $v_{\text{own}}$ $\leq$ 811, 600 $\leq$ $v_{\text{int}}$ $\leq$ 943}.
    \item \textit{30093 $\leq$ range $\leq$ 60760,  2.51 $\leq$ $\theta$ $\leq$ 2.81, -3.14 $\leq$ $\psi$ $\leq$ -3.13, 872 $\leq$ $v_{\text{own}}$ $\leq$ 1077, $v_{\text{int}}$ $=$ 0}.
    \item \textit{15624 $\leq$ range $\leq$ 28109, $\theta$ $=$ 2.82, $\psi$ $=$ -3.14, 1002 $\leq$ $v_{\text{own}}$ $\leq$ 1200, 585 $\leq$ $v_{\text{int}}$ $\leq$ 1009}.
    \item \textit{24392 $\leq$ range $\leq$ 39861,  $\theta$ $=$ 2.98, $\psi$ $=$ -3.14, 394 $\leq$ $v_{\text{own}}$ $\leq$ 646, 900 $\leq$ $v_{\text{int}}$ $\leq$ 1200}.
    \item \textit{30139 $\leq$ range $\leq$ 53232, 2.51 $\leq$ $\theta$ $\leq$ 2.83, -3.14 $\leq$ $\psi$ $\leq$ -3.13, 871 $\leq$ $v_{\text{own}}$ $\leq$ 1200, $v_{\text{int}}$ $=$ 0}.
    \item \textit{35747 $\leq$ range $\leq$ 48608, $\theta$ $=$ 1.57, $\psi$ $=$ -3.14, 862 $\leq$ $v_{\text{own}}$ $\leq$ 1164, 450 $\leq$ $v_{\text{int}}$ $\leq$ 736}.
    \item \textit{23991 $\leq$ range $\leq$ 42230, 2.522 $\leq$ $\theta$ $\leq$ 2.94, -3.14 $\leq$ $\psi$ $\leq$ -3.13, 100 $\leq$ $v_{\text{own}}$ $\leq$ 205, 900 $\leq$ $v_{\text{int}}$ $\leq$ 1200}.
\end{enumerate}

\subsection{Tables}
Statistics from our experiments are presented in Tables
\ref{tbl:inputinvarianttable}, \ref{tbl:interinvarianttable} and \ref{tbl:ACASXU_inter_invariants}.
\onecolumn
{\footnotesize
 \begin{longtable}{|p{0.10\textwidth}|p{0.45\textwidth}|p{0.05\textwidth}|}
 \caption{Input Properties for MNIST listing layers: nodes in layer and support.\ignoreme{ With No Visualizations}\label{tbl:inputinvarianttable}}\\
 \hline
 Pattern:Label & Layers:Nodes & Support \\ [0.5ex]
 \hline
 \endhead
 $\dpattern_1$: 0 & 1:0-9, 2:0-9 \ignoreme{0:0,1:0,2:0,3:0,4:0,5:1,6:0,7:1,8:0,9:1,10:0,11:1,12:0,13:1, 14:0,15:1,16:1,17:0,18:1,19:0 \ignoreme{0-19 & 0,0,0,0,0,1,0,1,0,1,0, 1,0,1,0,1,1,0,1,0}} & 1928 \\
 \hline
 $\dpattern_2$: 0 & 1:0-9, 2:0-7 \ignoreme{0:0,1:0,2:0,3:0,4:0,5:1,6:0,7:1,8:0,9:1,10:0,11:1,12:0, 13:1,14:0,15:1,16:1,17:0 \ignoreme{0-17 & 0,0,0,0,0,1,0,1,0,1,0, 1,0,1,0,1,1,0}} & 2010 \\
 \hline
 $\dpattern_3$: 0 & 1:0-9, 2:0-9 \ignoreme{0:1,1:0,2:0,3:1,4:0,5:1,6:0,7:1,8:0,9:1,10:0,11:1,12:0, 13:1,14:0,15:1,16:1,17:0,18:1,19:0 \ignoreme{0-19 & 1,0,0,1,0,1,0,1,0,1,0, 1,0,1,0,1,1,0,1,0}} & 217 \\
 \hline
 $\dpattern_4$: 1 & 1:0-9, 2:0-9 \ignoreme{ 0:1,1:1,2:0,3:0,4:1,5:1,6:0,7:0,8:0,9:1,10:0,11:0,12:0,13:0, 14:0,15:0,16:0,17:1,18:0,19:0 \ignoreme{0-19 & 1,1,0,0,1,1,0,0,0,1,0, 0,0,0,0,0,0,1,0,0}} & 758 \\
 \hline
 $\dpattern_5$: 1 & 1:0-9, 2:0-5 \ignoreme{0:0,1:1,2:1,3:0,4:1,5:0,6:0,7:0,8:0,9:0,10:0,11:0,12:1, 13:0,14:0,15:0 \ignoreme{0-15 & 0,1,1,0,1,0,0,0,0,0,0, 0,1,0,0,1}} & 2 \\
 \hline
 $\dpattern_6$: 1 & 1:0-9, 2:0-9, 3:0-9, 4:\{5\} \ignoreme{0:1,1:1,2:0,3:1,4:1,5:0,6:0,7:0,8:0,9:1,10:0,11:0,12:0, 13:0,14:0,15:0,16:0,17:1,18:0,19:0,20:0,21:1,22:1,23:0, 24:0,25:0,26:1,27:1,28:0,29:0,35:0 \ignoreme{0-29,35 & 1,1,0,1,1,0,0,0,0,1,0, 0,0,0,0,0,0,1,0,0,0,1, 1,0,0,0,1,1,0,0,0}} & 12 \\
 \hline
 $\dpattern_7$: 2 & 1:0-9, 2:\{2,3,4,5,8,9\}\ignoreme{0:0,1:1,2:0,3:0,4:0,5:1,6:0,7:0,8:0,9:1,14:0,15:1,19:0,12:1, 13:1,18:0 \ignoreme{0-9,14,15,19,12,13, 18 & 0,1,0,0,0,1,0,0,0,1,0, 1,0,1,1,0}} & 1338 \\
 \hline
 $\dpattern_8$: 2 & 1:0-9, 2:0-9, 3:0 \ignoreme{0:1,1:1,2:0,3:0,4:0,5:1,6:0,7:0,8:0,9:1,10:1,11:1,12:1, 13:1,14:0,15:1,16:1,17:1,18:0,19:1,20:0 \ignoreme{0-20 & 1,1,0,0,0,1,0,0,0,1,1, 1,1,1,0,1,1,1,0,1,0}} & 19 \\
 \hline
 $\dpattern_9$: 2 & 1:0-9, 2:0 \ignoreme{0:0,1:1,2:0,3:0,4:0,5:1,6:0,7:1,8:1,9:1,10:1 \ignoreme{0-10 & 0,1,0,0,0,1,0,1,1,1,1}} & 4 \\
 \hline
 $\dpattern_{10}$: 3 & 1:0-9, 2:0-9, 3:0-9, 4:\{5\} \ignoreme{0:1,1:1,2:1,3:0,4:1,5:1,6:0,7:0,8:0,9:1,10:0,11:1,12:1,13:0, 14:0,15:0,16:0,17:1,18:1,19:0,20:1,21:1,22:0,23:0,24:1,25:0, 26:0,27:1,28:0,29:1,35:1 \ignoreme{0-29,35 & 1,1,1,0,1,1,0,0,0,1,0, 1,1,0,0,0,0,1,1,0,1, 1,0,0,1,0,0,1,0,1,1}} & 2 \\
 \hline
 $\dpattern_{11}$: 3 & 1:0-9, 2:0-9, 3:\{3\} \ignoreme{0:1,1:1,2:1,3:1,4:0,5:1,6:0,7:1,8:0,9:1,10:0,11:1,12:1, 13:0,14:0,15:0,16:1,17:1,18:1,19:0,23:0 \ignoreme{0-19,23 & 1,1,1,1,0,1,0,1,0,1,0, 1,1,0,0,0,1,1,1,0,0}} & 52 \\
 \hline
 $\dpattern_{12}$: 4 & 1:0-9, 2:0-9, 3:0 \ignoreme{0:1,1:0,2:0,3:1,4:0,5:0,6:0,7:0,8:1,9:0,10:0,11:1,12:0, 13:0,14:1,15:1,16:0,17:0,18:1,19:0,20:0 \ignoreme{0-20 & 1,0,0,1,0,0,0,0,1,0,0, 1,0,0,1,1,0,0,1,0,0}} & 97 \\
 \hline
 $\dpattern_{13}$: 4 & 1:0-9, 2:0-9, 3:\{4\} \ignoreme{0:0,1:0,2:0,3:0,4:1,5:1,6:0,7:0,8:1,9:0,10:0,11:1,12:1, 13:1,14:0,15:0,16:1,17:1,18:0,19:0,24:0 \ignoreme{0-19,24 & 0,0,0,0,1,1,0,0,1,0,0, 1,1,1,0,0,1,1,0,0,0}} & 10 \\
 \hline
 $\dpattern_{14}$: 5 & 1:0-9, 2:0-9, 3:0-9, 4:0-9, 5:0-9, 6:0-1 \ignoreme{0:0,1:1,2:1,3:0,4:1,5:1,6:0,7:0,8:1,9:1,1:10,11:1,12:0,13:1, 14:0,15:0,16:1,17:1,18:1,19:1,20:1,21:1,22:0,23:1,24:1,25:0, 26:0,27:0,28:0,29:0,30:0,31:0,32:0,33:0,34:0,35:1,36:1,37:1, 38:1,39:1,40:1,41:0,42:1,43:1,44:0,45:0,46:0,47:0,48:1,49:0, 50:0,51:0 \ignoreme{0-51 & 0,1,1,0,1,1,0,0,1,1,1, 1,0,1,0,0,1,1,1,1,1,1, 0,1,1,0,0,0,0,0,0,0,0, 0,0,1,1,1,1,1,1,0,1,1, 0,0,0,0,1,0,0,0}} & 1 \\
 \hline
 $\dpattern_{15}$: 5 & 1:0-9, 2:0-9, 3:0-9, 4:0-9, 5:\{2\} \ignoreme{0:1,1:1,2:1,3:0,4:1,5:1,6:0,7:0,8:0,9:1,10:1,11:1,12:0, 13:0,14:0,15:0,16:0,17:1,18:0,19:0,20:1,21:1,22:0,23:1, 24:1,25:0,26:0,27:0,28:0,29:0,30:0,31:0,32:0,33:0,34:0, 35:0,36:1,37:1,38:1,39:1,42:1 \ignoreme{0-39,42 & 1,1,1,0,1,1,0,0,0,1,1, 1,0,0,0,0,0,1,0,0,1,1, 0,1,1,0,0,0,0,0,0,0,0, 0,0,0,1,1,1,1,1}} & 2 \\
 \hline
 $\dpattern_{16}$: 6 & 1:0-9, 2:\{0,5\} \ignoreme{0:0,1:0,2:0,3:0,4:0,5:0,6:0,7:0,8:0,9:1,15:1,10:0 \ignoreme{0-9,15,10 & 0,0,0,0,0,0,0,0,0,1,1 0}} & 748 \\
 \hline
 $\dpattern_{17}$: 6 & 1:0-9, 2:0 \ignoreme{0:0,1:0,2:0,3:0,4:0,5:0,6:0,7:0,8:0,9:0,10:0 \ignoreme{0-10 & 0,0,0,0,0,0,0,0,0,0,0}} & 3904 \\
 \hline
 $\dpattern_{18}$: 8 & 1:0-9, 2:\{0,2,4,5,8\} \ignoreme{0:1,1:0,2:0,3:0,4:0,5:1,6:0,7:0,8:0,9:1,12:0,14:0,10:0,18:1, 15:0 \ignoreme{0-9,12,14,10,18,15 & 1,0,0,0,0,1,0,0,0,1,0, 0,0,1,0}} & 358 \\
 \hline
 $\dpattern_{19}$: 8 & 1:0-9, 2:0-9, 3:0-9, 4:0-9, 5:0-9, 6:0-5 \ignoreme{0:1,1:1,2:0,3:1,4:1,5:0,6:0,7:0,8:0,9:1,10:0,11:1,12:0, 13:0,14:0,15:0,16:1,17:1,18:1,19:1,20:1,21:1,22:0,23:0, 24:0,25:0,26:0,27:1,28:0,29:0,30:0,31:0,32:0,33:0,34:0, 35:0,36:1,37:1,38:1,39:1,40:1,41:0,42:0,43:,44:0,45:0, 46:0,47:0,48:1,49:0,50:1,51:1,52:0,53:0,54:1,55:1  \ignoreme{0-55 & 1,1,0,1,1,0,0,0,0,1,0, 1,0,0,0,0,1,1,1,1,1,1, 0,0,0,0,0,1,0,0,0,0,0, 0,0,0,1,1,1,1,1,0,0,1, 0,0,0,0,1,0,1,1,0,0,1, 1}} & 3 \\
 \hline
 $\dpattern_{20}$: 9 & 1:0-9, 2:0-9, 3:0-9, 4:0-2 \ignoreme{0:1,1:0,2:0,3:1,4:0,5:0,6:1,7:0,8:1,9:1,10:0,11:1,12:0,13:0, 14:1,15:0,16:0,17:0,18:1,19:0,20:1,21:1,22:0,23:1,24:1,25:0, 26:1,27:1,28:0,29:1,30:0,31:1,32:0 \ignoreme{0-32 & 1,0,0,1,0,0,1,0,1,1,0, 1,0,0,1,0,0,0,1,0,1,1, 0,1,1,0,1,1,0,1,0,1,0,}} & 236 \\
 \hline
 $\dpattern_{21}$: 9 & 1:0-9, 2:0-9, 3:0-9, 4:0-9, 5:0-9, 6:0-9, 7:0-9, 8:0-9, 9:0-9 \ignoreme{0:1,1:0,2:1,3:1,4:0,5:0,6:1,7:0,8:1,9:1,10:0,11:1,12:0, 13:0,14:0,15:0,16:0,17:1,18:1,19:1,20:1,21:1,22:0,23:1, 24:1,25:0,26:1,27:1,28:0,29:1,30:0,31:1,32:0,33:0,34:0, 35:0,36:0,37:1,38:1,39:1,40:1,41:0,42:0,43:1,44:0,45:1, 46:0,47:1,48:0,49:0,50:0,51:0,52:1,53:1,54:0,55:0,56:1, 57:1,58:1,59:0,60:1,61:1,62:1,63:1,64:1,65:1,66:0,67:0, 68:0,69:1,70:1,71:1,72:0,73:1,74:1,75:0,76:0,77:1,78:1, 79:0,80:1,81:0,82:0,83:1,84:0,85:0,86:0,87:1,88:1,89:1  \ignoreme{0-89 & 1,0,1,1,0,0,1,0,1,1,0, 1,0,0,0,0,0,1,1,1,1,1, 0,1,1,0,1,1,0,1,0,1,0, 0,0,0,0,1,1,1,1,0,1,1, 0,1,0,1,0,0,0,0,1,1,0, 0,1,1,1,0,1,1,1,1,1,1, 0,0,0,1,1,1,0,1,1,0,0, 1,1,0,1,0,0,1,0,0,0,1, 1,1}} & 10 \\
 \hline
 $\dpattern_{22}$: 9 & 1:0-9, 2:0-9, 3:0-9, 4:0-9, 5:0-9, 6:0-9, 7:0-9, 8:0-9, 9:0-9, 10:0-9 \ignoreme{0:1,1:0,2:1,3:1,4:0,5:1,6:1,7:0,8:1,9:1,10:0,11:1,12:0, 13:1,14:0,15:0,16:0,17:1,18:1,19:0,20:1,21:1,22:0,23:1, 24:1,25:0,26:1,27:1,28:0,29:1,30:0,31:1,32:0,33:0,34:0, 35:1,36:0,37:1,38:1,39:1,40:0,41:0,42:1,43:1,44:0,45:1, 46:0,47:1,48:0,49:0,50:0,51:0,52:1,53:1,54:0,55:0,56:1, 57:0,58:1,59:0,60:1,61:1,62:1,63:1,64:1,65:1,66:0,67:0, 68:0,69:1,70:1,71:1,72:0,73:0,74:1,75:0,76:0,77:1,78:1, 79:0,80:1,81:0,82:0,83:1,84:0,85:0,86:0,87:1,88:1,89:1, 90:1,91:1,92:0,93:0,94:1,95:0,96:0,97:1, 98:1,99:0 \ignoreme{0-99 & 1,0,1,1,0,1,1,0,1,1,0, 1,0,1,0,0,0,1,1,0,1,1, 0,1,1,0,1,1,0,1,0,1,0, 0,0,1,0,1,1,1,0,0,1,1, 0,1,0,1,0,0,0,0,1,1,0, 0,1,0,1,0,1,1,1,1,1,1, 0,0,0,1,1,1,0,0,1,0,0, 1,1,0,1,0,0,1,0,0,0,1, 1,1,1,1,0,0,1,0,0,1,1, 0}} & 1 \\
 \hline
\end{longtable}
}

\begin{center}
{\footnotesize
 \begin{longtable}{|p{0.10\textwidth}|p{0.45\textwidth}|p{0.05\textwidth}|}
 \caption{Layer Properties for MNIST listing layer: nodes in layer and support. 
 \label{tbl:interinvarianttable}}\\
 \hline
 Pattern:Label & Layers:Nodes & Support \\ [0.5ex]
 \hline
 \endhead
 $\dpattern_{1}$: 6 & 1:0-9 \ignoreme{4:0,8:0,7:0,1:0,0:0,2:0,5:0,3:0,9:0,6:0 & \ignoreme{0,0,0,0,0,0,0,0,0,0}} & 3904 \\
 \hline
 $\dpattern_{2}$: 6 & 7:\{1-4, 7, 9\} \ignoreme{69:1,63:1,62:0,67:1,61:0,64:1 \ignoreme{& 1,1,0,1,0,1}} & 5145 \\
 \hline
 $\dpattern_{3}$: 4 & 6:\{0-2, 4-6, 8\} \ignoreme{56:1,54:1,52:1,51:0,50:1,55:0,58:0 \ignoreme{& 1,1,1,0,1,0,0}} & 3078 \\
 \hline
 $\dpattern_{4}$: 0 & 7:\{1-2, 4-5, 7, 9\} \ignoreme{69:1,62:0,67:0,64:0,65:1,61:0 \ignoreme{& 1,0,0,0,1,0}} & 5333 \\
 \hline
 $\dpattern_{5}$: 0 & 2:0-9, 3:0-7 \ignoreme{10:0,11:1,12:0,13:1,14:0,15:1,16:1,17:0,18:1,19:0,20:0, 21:1,22:0,23:0,24:0,25:1,26:1,27:0 \ignoreme{10-27 & 0,0,0,0,0,1,0,1,0,1,0, 1,0,1,0,1,1,0}} & 19962 \\
 \hline
 $\dpattern_{6}$: 3 & 9:\{0, 2-4, 6, 8-9\} \ignoreme{84:1,88:1,86:0,82:1,83:1,80:1,89:0 \ignoreme{& 1,1,0,1,1,1,0}} & 3402 \\
 \hline
 $\dpattern_{7}$: 5 & 10:\{0, 2, 4-5, 7-8\} \ignoreme{97:1,90:1,92:1,95:1,98:0,94:0 \ignoreme{& 1,1,1,1,0,0,}} & 3075 \\
 \hline
 $\dpattern_{8}$: 1 & 2:0-9, 3:0 \ignoreme{10:0,11:0,12:1,13:0,14:0,15:1,16:0,17:0,18:0,19:0,20:0 \ignoreme{10-20 & 0,0,1,0,0,1,0,0,0,0,0}} & 18735 \\
 \hline
 \end{longtable}
 }
\end{center}

\begin{center}
    {\footnotesize
    \begin{longtable}{|m{0.07\textwidth}|p{0.07\textwidth}|p{0.15\textwidth}|p{0.15\textwidth}|p{0.17\textwidth}|}
    \caption{ACASXU Layer Patterns, listing number of patterns, total support and the maximum support for a pattern.\label{tbl:ACASXU_inter_invariants}}\\
    \hline
    Layer & Label & Num of Patterns & Total Supp & MAX supp inv \\
    \hline
    \endhead
    5 & 0 & 834 & 2237734 & 109147 \\
    \hline
    5 & 1 & 776 & 3742 & 120 \\
    \hline
    5 & 2 & 1139 & 7744 & 1324 \\
    \hline
    5 & 3 & 1745 & 20059 & 2097 \\
    \hline
    5 & 4 & 1590 & 23580 & 2133 \\
    \hline
    4 & 0 & 1554 & 208136 & 25489 \\
    \hline
    4 & 1 & 1185 & 7338 & 732 \\
    \hline
    4 & 2 & 1272 & 7436 & 745 \\
    \hline
    4 & 3 & 2322 & 22880 & 1424 \\
    \hline
    4 & 4 & 2156 & 24565 & 2138 \\
    \hline
    3 & 0 & 3923 & 249771 & 26134 \\
    \hline
    3 & 1 & 1906 & 7387 & 210 \\
    \hline
    3 & 2 & 1866 & 6649 & 134 \\
    \hline
    3 & 3 & 3420 & 21902 & 945 \\
    \hline
    3 & 4 & 2932 & 20218 & 552 \\
    \hline
    2 & 0 & 1924 & 219149 & 51709 \\
    \hline
    2 & 1 & 734 & 4960 & 497 \\
    \hline
    2 & 2 & 819 & 4460 & 571 \\
    \hline
    2 & 3 & 1746 & 14487 & 1262 \\
    \hline
    2 & 4 & 1640 & 14571 & 1410 \\
    \hline
    1 & 0 & 2937 & 220395 & 32384 \\
    \hline
    1 & 1 & 1031 & 4422 & 265 \\
    \hline
    1 & 2 & 1123 & 3611 & 148 \\
    \hline
    1 & 3 & 2285 & 11756 & 311 \\
    \hline
    1 & 4 & 2112 & 11386 & 437 \\
    \hline
    \end{longtable}
    }
\end{center}

\end{document}